\documentclass[11pt, a4paper, goog]{google}

\usepackage[authoryear, sort&compress, round]{natbib}
\usepackage{tikz}
\usepackage{xcolor}
\usepackage{booktabs}
\usepackage{multirow}
\usetikzlibrary{shapes.geometric, arrows.meta, positioning, calc, fit}
\usepackage{soul}

\usepackage[colorinlistoftodos]{todonotes}
\keywords{machine learning, cryptography, number theory, scaling laws}

\uselogo{}

\title{Semiprime Factorization with Discrete Diffusion and Beam Search: Scaling Laws}

\correspondingauthor{nenadt@google.com}

    \author[*,1]{Brendan O'Donoghue}
    \author[*,1]{Nenad Toma\v{s}ev}
    \author[*,1]{Volodymyr Mnih}
    \author[*,1]{Timothy Nguyen}

    \author[1]{Dalin Guo}
    \author[1]{Arun Nair}
    \author[1]{Kenneth Yang}
    \author[1]{Alexandre Fr\'{e}chette}
    \author[1]{Jasper Xian}
    \author[1]{Jean Tarbouriech}
    \author[1]{Pavel Tyletski}
    \author[1]{Ingrid von Glehn}
    \author[1]{Johannes Bausch}
    \author[1]{Demis Hassabis}
    \author[1]{Raia Hadsell}

    \affil[*]{Equal contributions}
    \affil[1]{Google DeepMind}

\begin{abstract}
We demonstrate that transformer architectures trained with discrete denoising diffusion and decoded via guided test-time beam search can learn to recover the prime factors of semiprime integers directly from data. Our approach establishes a new frontier for learning-based semiprime factorization, successfully factoring 68-bit integers—a substantial advance over the prior 56-bit state of the art. While computationally less efficient than classical factoring algorithms, our model requires no domain-specific number-theoretic priors and learns purely from training examples. In practice, the diffusion model acts as a generative prior, identifying promising factor candidates where exact verification or lattice-based post-processing can complete the factorization. Furthermore, we derive empirical scaling laws relating approximation quality to training data volume, model size, and test-time computation. In contrast to language, problem size is the key driver of scaling behavior: achieving a fixed accuracy on a held-out test set requires compute that scales exponentially with bit length. Consequently, our findings pose no threat to security, as the projected compute cost required to factorize cryptographically relevant integers remains prohibitive.

\end{abstract}

\begin{document}

\maketitle

\section{Introduction}

Integer factorization is one of the most important open problems in computer science, cryptography, and algorithmic number theory. Given a composite integer $n$, the objective is to identify its prime factors such that $n=p_1 \cdot p_2 \dots p_k$. In the context of public-key cryptography, such as the Rivest--Shamir--Adleman (RSA) cryptosystem~\citep{rivest1978method}, the modulus $n$ is restricted to be a balanced semiprime---the product of two distinct, large, and randomly chosen prime factors $p$ and $q$ of equal bit-length. The security guarantees of RSA rely on the assumption that the inverse operation of recovering $p$ and $q$ from $n$ is intractable without additional knowledge.

The integer factorization problem is known to belong to the complexity classes NP (Nondeterministic Polynomial time) and co-NP~\citep{papadimitriou2003computational}, as any proposed solution can be verified in polynomial time. However, it is widely believed to be neither NP-complete nor contained within the polynomial-time class P, even though there is not yet a definitive proof that integer factorization on classical hardware requires exponential or super-polynomial time. Consequently, this problem has attracted substantial research interest. A discovery of an efficient classical factorization method would undermine the security of modern digital infrastructure, making the integer factorization problem not only a question of broader theoretical interest, but one of deep practical significance.

There is no shortage of increasingly sophisticated heuristics that have been proposed over a long period of time. Early naive approaches, such as trial division and Pollard's rho algorithm~\citep{pollalrd1975monte}, require exponential time. Therefore, these methods are inapplicable at the cryptographically relevant scale. The current state-of-the-art classical algorithm is the General Number Field Sieve (GNFS)~\citep{lenstra1990number, lenstra1993development, buhler2006factoring}, which was introduced as an evolution of the Quadratic Sieve (QS)~\citep{pomerance1985quadratic}. GNFS involves constructing two monic irreducible polynomials that share a common root modulo $n$, and the search is performed across smooth numbers over complex algebraic number fields. The time complexity of GNFS is sub-exponential, namely $$L_n[1/3, c]=\exp\left((c+o(1))(\ln n)^{1/3}(\ln \ln n)^{2/3}\right)$$
where $c=(64/9)^{1/3} \approx 1.923$. While GNFS outperforms other methods, it remains super-polynomial in the bit-length of $n$. Consequently, an 896-bit RSA cryptographic modulus currently remains the largest publicly acknowledged number to have been factored, which required substantial compute~\citep{weis2026rsa896}.

Subsequently, Peter Shor introduced a quantum algorithm capable of efficiently solving integer factorization presuming access to future fault-tolerant quantum hardware~\citep{shor1995polynomial, monz2016realization}. Shor's algorithm leverages quantum superposition and the Quantum Fourier Transform (QFT) to solve the underlying period-finding problem, reducing the factorization time complexity to a polynomial bound of $\mathcal{O}((\log n)^3)$. This places integer factorization within the Bounded-error Quantum Polynomial time class (BQP)~\citep{aaronson2010bqp}. While such quantum computers remain far in the future, this finding has catalyzed the search for post-quantum cryptography.

Unsurprisingly, semiprime factorization has also been approached with machine learning methods, and more recently deep learning~\citep{jansen2005neural, nene2022machine, blake2023integer}. This has proven to be challenging, given the discrete combinatorial nature of the problem. Minor changes in activations may translate into severe violations of the imposed constraints. Recently, machine learning approaches have moved away from one-shot prediction to generative models. By formulating integer factorization as a search over a constrained discrete space, it is possible to achieve effective approximations at smaller scales. Specifically, the introduction of discrete denoising diffusion models~\citep{austin2021structured} has shown promise.

Here, we contribute an empirical study of the scaling laws for semiprime factorization via discrete diffusion, both with and without subsequent test-time search and post-processing. These scaling laws enable us to characterize the extent to which machine learning can hope to tackle this problem, and extrapolate the findings to project how much time and compute would be required. We conclude that this does not present a risk to the current RSA protocols. Finally, at the 68-bit ($k=34$) scale, we evaluate a 12B-parameter diffusion model paired with Coppersmith lattice post-processing across $15{,}523$ held-out 68-bit semiprimes, recovering factors on a single test instance ($1$ candidate per instance) and analyzing the full MSB prefix-match distribution to show that the model captures coarse multiplicative magnitude bounds ($\approx 3\times$ lift over uniform random bit-guessing) while decaying to near-chance accuracy on deeper factor bits.

\section{Related Work}

\subsection{Factorization}

The most promising classical factorization algorithms utilize the principle of identifying \emph{smooth numbers}, which are integers with bounded prime factors~\citep{dixon1981asymptotically}. The Quadratic Sieve (QS)~\citep{pomerance1985quadratic, pomerance1996tale} aims to discover integers $x$ and $y$ such that $x^2 \equiv y^2 \pmod n$ while ensuring $x \not\equiv \pm y \pmod n$, so that $\gcd(x-y, n)$ would then yield a non-trivial factor of $n$.

The General Number Field Sieve (GNFS)~\citep{buhler2006factoring} replaces the quadratic polynomials utilized in the QS with complex, higher-degree polynomials over algebraic number fields. The problem then becomes finding relations among elements within the ring of integers of a specific number field. During the sieve phase, many smaller integers are then factored to identify smooth relations, which can be achieved via Lenstra's elliptic-curve factorization method~\citep{lenstra1987factoring}.

Fermat's factorization method presents an alternative approach, which is centered around the idea of representing an odd composite integer $n$ as the difference of two distinct squares: $n=x^2-y^2=(x-y)(x+y)$. This is highly efficient if the prime factors $p$ and $q$ are close to the square root of $n$, within the asymptotic bound of $\mathcal{O}(n^{1/4})$~\citep{lehman1974factoring}. In 1895, Lawrence extended Fermat's algorithm for scenarios where $p/q$ is not close to 1~\citep{lawrence1895factorisation, hittmeir2021time}. It was shown that if one can deduce an approximation ratio $u/v \approx p/q$, one can then construct an augmented multiplier $m=uvn$ and use Fermat's method to factor $m$ quickly. Recently, there has been interest in utilizing Lawrence's extension to reframe the factorization problem into a binary classification or regression problem~\citep{blake2023integer} for deep learning. In this framing, neural networks are trained to estimate the $u/v$ ratio. While this framing is an easier output specification for neural networks, the performance has not been promising.

\subsection{Neural Arithmetic}

Using neural networks for integer factorization has proven to be challenging, which is due to highly non-smooth optimization landscapes~\citep{velivckovic2019neural} and the hard combinatorial nature of the problem. To overcome these obstacles, Vector Symbolic Architectures (VSA)~\citep{gayler2004vector} represent complex sub-symbolic data structures as high-dimensional hypervectors~\citep{kanerva2009hyperdimensional}. Fractional Power Encoding (FPE)~\citep{plate1995holographic, frady2022computing} is frequently used within the VSA framework, involving repeated exponentiation of a base hypervector and binding these to different scalar quantities.

Resonator networks~\citep{frady2020resonator, renner2024neuromorphic} maintain several independent groups of hypervector estimates, $\hat{a}(t)$ and $\hat{b}(t)$, in superposition, and update them iteratively based on the codebook matrices of possible factor representations. This dynamical system converges to a stable equilibrium state once the correct integer factors are isolated~\citep{kleyko2022integer}.

\subsection{Diffusion Models}

Denoising diffusion models were initially developed primarily for continuous domains~\citep{sohl2015deep, ho2020denoising, croitoru2023diffusion, cao2024survey} and have seen extensive applications in image generation~\citep{ulhaq2022efficient, croitoru2023diffusion}. Gaussian noise is added in the forward pass and then iteratively removed in the learned neural inverse process. Discrete problems were initially approached mainly via standard autoregressive approaches, yet these approaches face difficulties in highly structured problems as an early error in the sequence would invalidate the rest of the solution. Discrete diffusion models help overcome this obstacle~\citep{austin2021structured}.

Discrete diffusion models are a good fit for highly structured problems as they couple parallel generation and iterative global refinement. These models are often defined with a Markovian transition matrix, rather than Gaussian noise~\citep{hoogeboom2021argmax}. Recent theoretical findings, tying the negative Evidence Lower Bound (ELBO) to the data entropy and the path Kullback-Leibler (KL) divergence, suggest that the diffusion loss naturally leads to the model correctly capturing the conditional expectation of the reverse denoising steps~\citep{austin2021structured}. Discrete diffusion has shown great promise in structured domains like graph generation~\citep{vignac2022digress, haefeli2022diffusion, limnios2023sagess, chen2023efficient}.

Recently, discrete diffusion has been experimented with for integer factorization~\citep{freivalds2023discrete}. The idea is to learn a mapping that iteratively corrects errors within partially correct solutions. The neural network is trained to output a clean sample $x_0=\mu(x_t, \bar{\alpha}_t)$, which is tightly conditioned on the product sequence. The approximation utilizes the full probabilistic information that is retained between steps:

$$\tilde{\theta}=[\alpha_t x_t+(1-\alpha_t)/K] \odot [\bar{\alpha}_{t-1} x_0+(1-\bar{\alpha}_{t-1})/K]$$

With this approach, the model was able to recover prime factors for numbers up to 56 bits in length~\citep{freivalds2023discrete}. This work also highlighted an inverse exponential relationship between training compute allocation and inference cost: At a specific projected success rate, higher investment in pre-training has been shown to lead to an exponential decrease in the number of discrete sampling steps required.

\section{Method}
\label{sec:method}

\subsection{Data Representation}

We formulate the semiprime factorization problem, $n = p \times q$, as a conditional sequence generation task. Given an integer $n$, the objective is to generate the prime factors $p$ and $q$. We represent all integers as binary strings. Assuming the prime factors $p$ and $q$ are bounded by $2^k$, $k$ being the maximum number of factor bits, the binary representations of $p$ and $q$ are padded with leading zeros towards a fixed length $k$. Similarly, $n$ occupies $2k$ bits. The employed ordering goes from the Most Significant Bit (MSB) to the Least Significant Bit (LSB).

The input to the model is obtained by concatenating the bit strings for $n$, $p$, $q$, resulting in a total sequence length of $4k$. The vocabulary consists of binary tokens $\{0, 1\}$, alongside designated mask and padding tokens. During generation, the $2k$ bits corresponding to $n$ are unmasked and act as a fixed prefix. The model is tasked with inferring the subsequent $2k$ masked bits corresponding to $p$ and $q$.

\subsection{Data Generation}

Training examples are generated on the fly using deterministic pseudorandom streams, while excluding pre-generated held-out evaluation sets to prevent pair-level train/test leakage. At initialization, we pre-generate fixed held-out validation and test splits of $1{,}024$ semiprimes each (with a larger $100{,}000$-example held-out test set generated specifically for the 68-bit factorization evaluation in Section~\ref{sec:factorization_68bit}); during on-the-fly training example generation, any sampled pair whose product $n$ appears in the held-out evaluation splits is rejected and resampled. To sample a factor pair, we draw two random integers uniformly from $[0, 2^k - 1]$ (without forcing the leading $k$-th bit to $1$, so factors have bit-length at most $k$ and are zero-padded to length $k$), advance each to the next prime via \texttt{gmpy2.next\_prime} while rejecting primes $\ge 2^k$ or equal factors ($p = q$), sort them so that $p > q$ to break permutation symmetry, and set $n = p \cdot q$.

Note that this holdout protocol enforces \emph{pair-level} disjointness ($N_{\mathrm{train}} \cap (N_{\mathrm{val}} \cup N_{\mathrm{test}}) = \emptyset$), ensuring the model never sees a held-out semiprime $n$ during training, while drawing training and test pairs from the same underlying pool of primes $< 2^k$ ($P_{\mathrm{train}} = P_{\mathrm{test}}$). Let $\pi(n)$ denote the number of primes less than or equal to $n$. For small bit-lengths ($k \le 19$), a $100{,}000$-step training run at batch size $2^{14} = 16{,}384$ ($1.64 \times 10^9$ examples) makes multiple passes over the $\binom{\pi(2^k)}{2}$ non-held-out training pairs (e.g., $\approx 266$ passes over $\approx 6.17 \times 10^6$ pairs at $k=15$, and $\approx 77$ passes over $\approx 2.14 \times 10^7$ pairs at $k=16$) and exposes each individual prime factor thousands of times (e.g., $\approx 75{,}500$ occurrences per prime across $\pi(2^{19}) = 43{,}390$ primes at $k=19$). We discuss the implications of prime-vocabulary memorization versus pair-level generalization in Section~\ref{sec:conclusion}.

\subsection{Model Architecture and Self-Conditioned Training}

We approach semiprime factorization as a machine learning problem, framed as discrete diffusion over binary token sequences, utilising a bidirectional transformer~\citep{vaswani2023attentionneed} denoiser. During training, we use a noise schedule to corrupt the factor bits by replacing a dynamically sampled proportion $\eta \in [0.02, 1.0]$ of their tokens with the mask token, and train the model to reconstruct the clean binary values across all $2k$ factor positions.

The spatial structure of the forward noising process is critical for learning multi-bit arithmetic dependencies, leading us to adopt \emph{contiguous blockwise masking} rather than independent bitwise masking. For each training example we sample a noise level $\eta$ from the triangular density $p(\eta) \propto 2\eta$ on $[0.02, 1.0]$ (biasing towards higher noise levels while applying inverse-density importance weighting $w = 1/(2\eta)$ so the diffusion objective remains unbiased; see Appendix~\ref{app:training_rollout_and_loss}) and mask a contiguous block of length $M \approx \eta k$ in $p$ and in $q$, while leaving the $2k$ prefix bits of $n$ unmasked. Masking contiguous spans forces the denoiser to resolve long-range carry propagation and global multiplicative constraints rather than relying on local bit interpolation.

To allow the discrete denoiser to refine its factor estimates across iterations without losing continuous uncertainty information when discretizing to hard tokens $\{0, 1\}$, we incorporate a gated feed-forward \emph{self-conditioning} module~\citep{chen2023analogbitsgeneratingdiscrete}. Rather than feeding only discrete tokens at each step, the model maps its previous step's vocabulary logits $\boldsymbol{\ell}$ into a continuous probability-weighted embedding $\mathbf{S}(\boldsymbol{\ell}) = \sqrt{d}\,\mathrm{softmax}(\boldsymbol{\ell})\hat{\mathbf{E}}$ over the shared normalized token embedding table $\hat{\mathbf{E}}$, transforms $\mathbf{S}(\boldsymbol{\ell})$ through a zero-bias gated GeLU feed-forward network $\mathbf{G}_{\text{SC}}$, and adds the result residually to the discrete canvas token embeddings $\mathbf{C}$ prior to the first transformer layer (Appendix~\ref{app:self_conditioning}).

Training proceeds via a \emph{two-pass self-conditioned rollout} on each minibatch (detailed in Appendix~\ref{app:training_rollout_and_loss}):
\begin{enumerate}
    \item \textbf{Pass 1 (Unconditioned Proposal):} The denoiser is evaluated on the block-masked canvas $\tilde{\mathbf{x}}$ with an all-zero self-conditioning tensor $\mathbf{S}^{(0)} = \mathbf{0}$ (and $\mathrm{stop\_gradient}$) to produce first-pass logits $\boldsymbol{\ell}^{(1)}$ and soft self-conditioning embeddings $\mathbf{S}^{(1)} = \mathrm{stop\_gradient}(\mathbf{S}(\boldsymbol{\ell}^{(1)}))$.
    \item \textbf{Concave Canvas Resampling and Pass 2 (Self-Conditioned Update):} During iterative inference, the denoiser encounters canvases containing a mixture of masked spans and potentially erroneous discrete bits committed at earlier steps. To match this distribution during training, we sample discrete tokens $\hat{\mathbf{x}}^{(1)} \sim \mathrm{Categorical}(\boldsymbol{\ell}^{(1)})$ and overwrite each factor position of $\tilde{\mathbf{x}}$ (both masked and initially unmasked) independently with probability $p_{\text{replace}}(\eta) = 4\rho\eta(1-\eta)$ ($\rho = 0.5$; Eq.~\ref{eq:concave_replace}), forming a resampled canvas $\tilde{\mathbf{x}}^{(2)}$. On Pass 2, $\mathbf{S}^{(1)}$ is retained with probability $p_{\text{SC}} = 0.99$ and zeroed out with probability $1 - p_{\text{SC}} = 0.01$ so that a single parameter update per step trains the model predominantly in the self-conditioned regime while preserving calibration for the initial unconditioned step ($\mathbf{S}^{(0)} = \mathbf{0}$). The training objective computes the importance-weighted cross-entropy between the Pass-2 predicted logits and the ground-truth binary tokens across all $2k$ factor positions of $p$ and $q$ (Eq.~\ref{eq:ce_loss}), training the denoiser both to inpaint masked blocks and to correct erroneous discrete tokens committed on unmasked positions.
\end{enumerate}

\subsection{Test-Time Denoising and Guided Search}
\label{sec:test_time_search}

At inference time, we consider two distinct decoding procedures that trade off single-trajectory iterative refinement against structured tree search over partial factor prefixes (full mathematical details and a step-by-step worked walkthrough are provided in Appendix~\ref{app:beam_search_oracle_sc}).

\subsubsection{Non-Monotonic Ancestral Block-Diffusion Sampling}
\label{sec:ancestral_sampling_main}

Standard ancestral sampling generates a single candidate factorization over $T$ discrete steps ($t = 0, \dots, T-1$) following a linearly decaying noise schedule $\eta_{t+1} = 1 - \frac{t+1}{T}$, starting from a fully masked factor canvas ($\eta_0 = 1$) and $\mathbf{S}^{(0)} = \mathbf{0}$. Unlike absorptive discrete diffusion samplers that monotonically unmask and freeze one token at a time, our ancestral sampler performs \emph{non-monotonic predict-and-remask block refinement}. At each step $t$:
\begin{enumerate}
    \item The denoiser predicts clean binary logits $\boldsymbol{\ell}^{(t)}$ across all $2k$ factor bits simultaneously.
    \item The temperature-scaled binary probabilities are converted into the continuous self-conditioning embedding $\mathbf{S}(\tilde{\boldsymbol{\ell}}^{(t)})$ and passed to step $t+1$.
    \item A complete clean discrete guess $\hat{\mathbf{x}}_0^{(t)} \sim \mathrm{Categorical}(\tilde{\boldsymbol{\ell}}^{(t)})$ is sampled for all $2k$ bits of $p$ and $q$, and two \emph{fresh, independently positioned} contiguous blocks of width $M_t = \max(\lfloor \eta_{t+1} k \rfloor, 1)$ are re-masked at newly sampled random offsets in $p$ and $q$.
\end{enumerate}
Because the masked window slides randomly across $p$ and $q$ at every step while continuous beliefs accumulate in $\mathbf{S}(\tilde{\boldsymbol{\ell}}^{(t)})$, running $T \gg 2k$ steps (e.g., $T = 1{,}024$ to $16{,}384$ for $2k = 30$) acts as annealed blocked-Gibbs refinement: bits outside the current mask that were guessed inconsistently at earlier steps can be re-masked and corrected on subsequent steps once neighbouring bits settle. No intermediate arithmetic check (whether $\hat{p} \cdot \hat{q} = n$) is performed during the $T$ steps; verification occurs only once after step $T-1$.

\subsubsection{Guided Beam Search with Inner Mini-Denoising and Oracle Self-Conditioning}
\label{sec:beam_search_main}

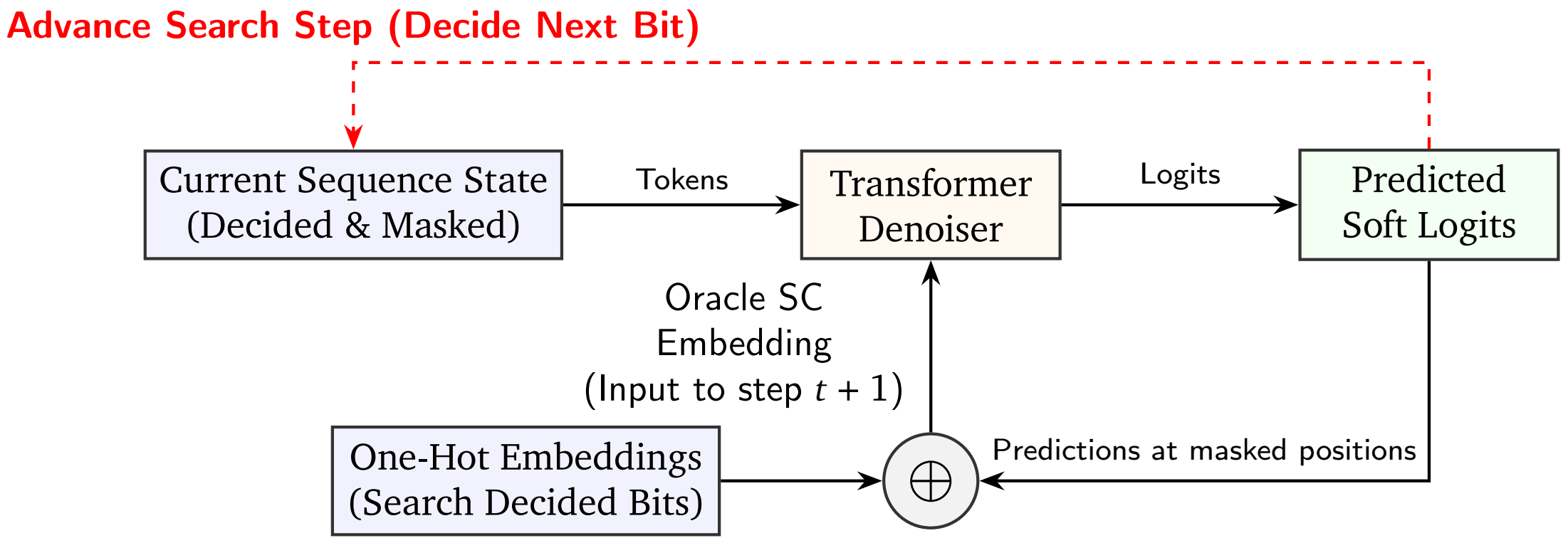
\begin{figure*}[htbp]
\centering
\begin{tikzpicture}[
    block/.style={rectangle, draw=black!80, fill=white, thick, minimum width=2.4cm, minimum height=1cm, align=center, font=\small},
    mux/.style={circle, draw=black!80, fill=gray!10, thick, inner sep=3pt, font=\small},
    arrow/.style={-Stealth, thick},
    labelstyle/.style={font=\small\sffamily, align=center}
]

    \node[block, fill=blue!5] (input) {Current Sequence State\\(Decided \& Masked)};
    \node[block, right=2.2cm of input, fill=orange!5] (denoiser) {Transformer\\Denoiser};
    \node[block, right=2.2cm of denoiser, fill=green!5] (logits) {Predicted\\Soft Logits};

    \draw[arrow] (input) -- (denoiser) node[midway, above, font=\scriptsize\sffamily] {Tokens};
    \draw[arrow] (denoiser) -- (logits) node[midway, above, font=\scriptsize\sffamily] {Logits};

    \node[mux, below=1.6cm of denoiser] (sc_merge) {$\bigoplus$};

    \node[block, fill=blue!5] (one_hot) [left=1.5cm of sc_merge] {One-Hot Embeddings\\(Search Decided Bits)};
    \draw[arrow] (one_hot) -- (sc_merge);

    \path [draw, arrow] (logits.south) |- (sc_merge.east)
        node[pos=0.75, above, font=\scriptsize\sffamily] {Predictions at masked positions};

    \draw[arrow] (sc_merge.north) -- (denoiser.south)
        node[midway, left=0.1cm, labelstyle] {Oracle SC\\Embedding\\(Input to step $t+1$)};

    \path [draw, arrow, dashed, red] (logits.north) -- ++(0,0.8) -| (input.north)
        node[pos=0.5, above, font=\small\sffamily\bfseries] {Advance Search Step (Decide Next Bit)};

\end{tikzpicture}
\caption{The oracle self-conditioning iteration. During the inner mini-denoising loop, decided bits are mapped to one-hot target embeddings, while remaining masked positions receive soft logits. Both signals are merged to form the conditioning input for the subsequent denoising step.}
\label{fig:oracle_sc}
\end{figure*}

\begin{figure*}[htbp]
\centering
\resizebox{0.95\linewidth}{!}{%
\begin{tikzpicture}[
    level 1/.style={sibling distance=9.0cm, level distance=1.5cm},
    level 2/.style={sibling distance=4.5cm, level distance=1.5cm},
    node_valid/.style={circle, draw=green!80!black, fill=green!10, thick, minimum size=8mm, font=\small},
    node_pruned/.style={circle, draw=red!80, fill=red!10, thick, minimum size=8mm, inner sep=0pt, font=\small},
    node_root/.style={rectangle, draw=blue!80, fill=blue!10, thick, rounded corners, minimum size=9mm, font=\small},
    arrow/.style={-Stealth, thick},
    labelstyle/.style={font=\small\sffamily\itshape, anchor=west}
]
    \coordinate (left_bound) at (-9.5, 0);
    \node[node_root] (root) at (0,0) {$p_{\emptyset}, q_{\emptyset}$}
        child {node[node_valid] (l) {$p_0=0$}
            child {node[node_pruned] (ll) {$q_0=0$}
                edge from parent[arrow] node[left=0.15cm, font=\scriptsize\sffamily] {expand $q$}
            }
            child {node[node_valid] (lr) {$q_0=1$}
                edge from parent[arrow]
            }
            edge from parent[arrow] node[left=0.15cm, font=\scriptsize\sffamily] {expand $p$}
        }
        child {node[node_valid] (r) {$p_0=1$}
            child {node[node_valid] (rl) {$q_0=0$}
                edge from parent[arrow]
            }
            child {node[node_valid] (rr) {$q_0=1$}
                edge from parent[arrow]
            }
            edge from parent[arrow]
        };
    \node[right=0.25cm of ll, text=red, font=\footnotesize\sffamily\bfseries] {Pruned: Infeasible};

    \draw[red, thick] ($(ll.north west)+(0.1,-0.1)$) -- ($(ll.south east)+(-0.1,0.1)$);
    \draw[red, thick] ($(ll.south west)+(0.1,0.1)$) -- ($(ll.north east)+(-0.1,-0.1)$);

    \node[labelstyle] at (left_bound) {Step 0};
    \node[labelstyle] at (left_bound |- l) {Step 1};
    \node[labelstyle] at (left_bound |- ll) {Step 2};
    \draw[blue!40, dashed, thick, rounded corners]
        ($(ll.west)+(-0.3,-0.3)$) rectangle ($(rr.east)+(0.3,0.3)$);

    \node[below=0.4cm of ll.south east, anchor=north, font=\small\sffamily\bfseries, text=blue!80!black]
        {Beam Selection: Keep Top-$B$ Candidates by Probability};
\end{tikzpicture}%
}
\caption{Beam search expansion and feasibility pruning. At each step, valid states in the active beam are expanded. If a partial assignment violates the arithmetic constraints, it is pruned. The remaining feasible states are ranked, and the top $B$ paths are retained in the beam.}
\label{fig:beam_search_pruning}
\end{figure*}
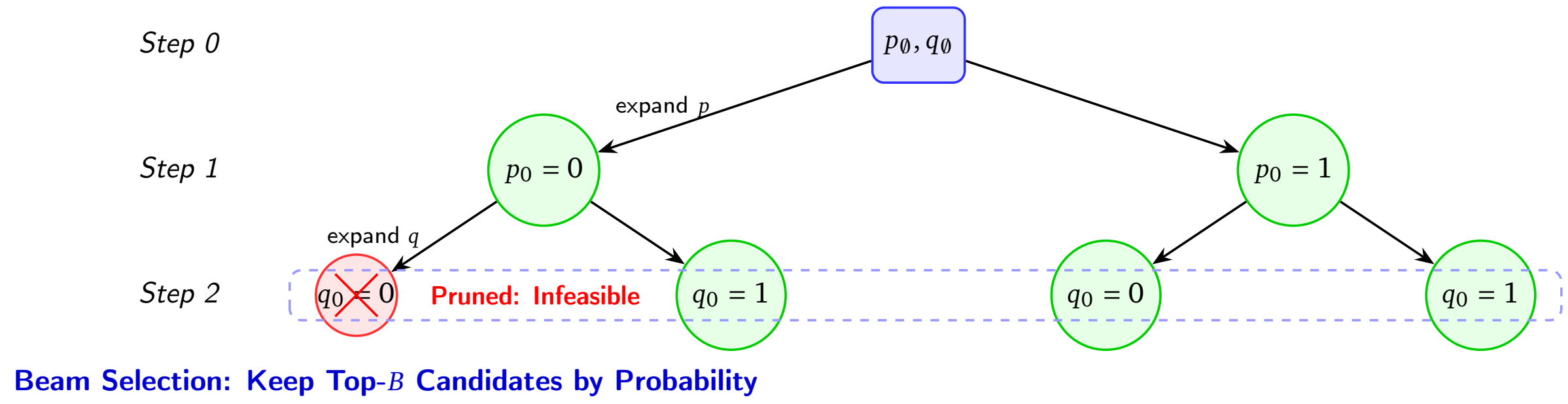

While non-monotonic ancestral sampling can revise earlier bit guesses, it still evolves a single stochastic trajectory without explicit arithmetic verification until the final step. To prevent early bit errors from derailing generation and to exploit deterministic arithmetic constraints during decoding, we replace single-trajectory generation with a two-level guided search procedure over the binary factor tokens (Figures~\ref{fig:oracle_sc} and~\ref{fig:beam_search_pruning}). While we had initially explored Monte Carlo Tree Search (MCTS) with PUCT selection~\citep{silver2017masteringchessshogiselfplay}, we empirically found it to be less effective than maintaining a synchronized frontier via beam search~\citep{lowerre1976harpy, lys2026d5p4partitiondeterminantalpoint} augmented with iterative inner mini-denoising.

Our procedure couples an outer beam search that commits to discrete factor bits one by one with an inner mini-denoising loop that uses the diffusion model—via a mechanism we term \textit{oracle self-conditioning}—to calibrate the branching probabilities at each search depth:
\begin{itemize}
    \item \textbf{Outer Loop (MSB-to-LSB Beam Search):} The outer search operates sequentially over the $2k$ factor positions ($i = 0, \dots, 2k-1$) by alternating between the MSBs of $p$ (even steps $i = 2r$, deciding bit $p_r$) and the MSBs of $q$ (odd steps $i = 2r+1$, deciding bit $q_r$). At outer depth $i$, each hypothesis node in the active beam of width $B$ partitions the $2k$ factor positions into a subset of already committed (\emph{decided}) MSB positions $\mathcal{D}_i$ with fixed binary assignments $x_j^* \in \{0, 1\}$ ($j \in \mathcal{D}_i$) and a subset of remaining \emph{undecided} masked positions $\mathcal{U}_i$.
    \item \textbf{Inner Loop (Mini-Denoising with Oracle Self-Conditioning):} Before expanding a node at outer depth $i$ to decide the next target MSB $t_i \in \mathcal{U}_i$, we run a short inner rollout of $T_{\text{mini}} = 5$ recurrent denoising passes ($m = 0, \dots, T_{\text{mini}}-1$) to let the diffusion model refine its beliefs over the undecided bits $\mathcal{U}_i$ conditioned on the committed prefix $\mathcal{D}_i$. In standard diffusion self-conditioning, the model feeds back its own soft predictions across all factor positions. In \textit{oracle self-conditioning} (Figure~\ref{fig:oracle_sc}), any bit $j \in \mathcal{D}_i$ already committed by the outer beam search is treated as ground-truth (``oracle'') knowledge during the inner rollout: decided positions $j \in \mathcal{D}_i$ are clamped in the self-conditioning input to their exact one-hot token embeddings $\sqrt{d}\,\mathbf{e}_{x_j^*}^\top\hat{\mathbf{E}}$~\citep{chen2023analogbitsgeneratingdiscrete}, while undecided masked positions $u \in \mathcal{U}_i$ continue to refine soft probability-weighted embeddings $\sqrt{d}\,\mathrm{softmax}(\tilde{\boldsymbol{\ell}}_u^{(m)})\hat{\mathbf{E}}$ across the $T_{\text{mini}}$ inner steps.
\end{itemize}

Even though the output logits at already-decided positions $j \in \mathcal{D}_i$ are discarded after the inner loop (since only the logit at the next undecided MSB position $t_i$ is read to expand the tree), clamping $j \in \mathcal{D}_i$ to one-hot self-conditioning embeddings is essential because the self-conditioning vector is added to the input canvas embeddings \emph{before} entering the bidirectional transformer. Through all-to-all self-attention across all layers, every undecided masked position $u \in \mathcal{U}_i$ attends directly to the keys and values at decided positions $j \in \mathcal{D}_i$. Injecting the one-hot embedding at $j \in \mathcal{D}_i$ both matches the Pass-2 training distribution on unmasked factor tokens and prevents internal conflict between the discrete canvas token $x_j^*$ and the model's unconditioned prior when exploring lower-probability branches of the beam (see Appendix~\ref{app:beam_search_oracle_sc}).

Putting the outer and inner loops together, the guided search proceeds over outer steps $i = 0, \dots, 2k-1$ across a fixed-size frontier of $B$ hypothesis slots as follows (detailed in Appendix~\ref{app:beam_search_oracle_sc}):

\begin{enumerate}
    \item \textbf{Initialization:} At outer step $i=0$, the frontier is initialized with $B$ slots where only slot $0$ is marked active ($\text{score} = 0.0, \text{valid} = \text{True}$) with $n$ unmasked and all $2k$ bits of $p$ and $q$ set to the mask token ($\mathcal{D}_0 = \emptyset$), while slots $1, \dots, B-1$ are marked inactive ($\text{score} = -\infty, \text{valid} = \text{False}$) to prevent duplicate clones at the root. Thus, the number of active hypotheses grows as $1 \to 2 \to 4 \to \dots \to B$.
    \item \textbf{Expansion via Inner Mini-Denoising ($T_{\text{mini}}$ steps):} At outer step $i$, for each active candidate in the beam, we execute the $T_{\text{mini}} = 5$ inner mini-denoising loop ($m = 0, \dots, T_{\text{mini}}-1$). Unlike ancestral sampling, the discrete canvas $\tilde{\mathbf{x}}$ is held \emph{completely fixed} throughout the inner loop (decided positions $j \in \mathcal{D}_i$ retain their committed tokens $x_j^* \in \{0,1\}$, undecided positions $u \in \mathcal{U}_i$ remain masked, and no re-masking occurs); only the continuous self-conditioning tensor $\mathbf{S}^{(m)}$ evolves from $\mathbf{S}^{(0)} = \mathbf{0}$ by repeatedly merging the fixed one-hot oracle embeddings on $\mathcal{D}_i$ with the refining soft predictions on $\mathcal{U}_i$ (Figure~\ref{fig:oracle_sc}). After $T_{\text{mini}}$ inner passes, the calibrated binary log-probabilities $\log \mathrm{softmax}(\boldsymbol{\ell}_{t_i}^{(T_{\text{mini}}-1)})$ at the target MSB position $t_i$ are used to expand each active parent into two child branches ($x_{t_i}^* \in \{0, 1\}$), yielding up to $2B$ candidate states.
    \item \textbf{Arithmetic Interval Feasibility Pruning:} Because bit expansion proceeds from MSB to LSB, any partial assignment of prefixes $p_{\le r}$ and $q_{\le r'}$ defines deterministic integer lower and upper bounds on each factor by setting all undecided lower bits in $\mathcal{U}_{i+1}$ to $0$ ($p_{\min}, q_{\min}$) or $1$ ($p_{\max}, q_{\max}$), respectively. Any child branch whose target semiprime falls outside the feasible product interval, i.e., $n \notin [p_{\min} q_{\min}, \, p_{\max} q_{\max}]$, is immediately pruned by setting its score to $-\infty$ (Figure~\ref{fig:beam_search_pruning}).
    \item \textbf{Top-$B$ Selection:} The remaining feasible children across all $2B$ candidates are ranked by their cumulative log-probabilities, and the top $B$ candidates are retained to form the active frontier for outer step $i+1$.
\end{enumerate}

At the final depth $i = 2k-1$ ($\mathcal{U}_{2k} = \emptyset$), $p_{\min} = p_{\max} = \hat{p}$ and $q_{\min} = q_{\max} = \hat{q}$, so interval feasibility pruning reduces to exact multiplicative verification $\hat{p} \cdot \hat{q} = n$. If all branches become infeasible prior to reaching the final depth, the search terminates early and returns the highest-log-probability partial MSB prefix for lattice-based post-processing.

\subsection{Post-Processing}
\label{sec:post_processing}

Optionally, we perform post-processing of the ML predictions via Coppersmith's algorithm~\citep{coppersmith1996finding}. Based on the Lenstra–Lenstra–Lovász (LLL) lattice basis reduction algorithm~\citep{lenstra1982factoring}, Coppersmith's method can efficiently find small roots of modular polynomial equations. In the context of semiprime factoring, Coppersmith's theorem states that if at least half of the MSBs of either prime factor are known, the remaining unknown bits can be recovered in polynomial time.

Consequently, for a factorization to be successful, it is not strictly necessary for either the model or the test-time search to generate a fully accurate factorization, if the generated candidate is approximately correct with a sufficient degree of accuracy. Coppersmith's method can complement the ML approaches to deterministically resolve the remaining bits.

\subsection{Test-Time Scaling Laws}

Given that scaling test-time compute is now seen as one of the primary directions for enhancing complex problem-solving capabilities in AI models~\citep{snell2024scalingllmtesttimecompute, chen2026provable}, our focus here was to not only develop scaling laws for discrete diffusion models in pretraining when attempting to solve the integer factorization problem, but perhaps even more importantly, understand the test-time scaling properties and the conditions under which beam search can help improve the factorization success rate at different levels of complexity, and for different levels of quality of the underlying diffusion model. Test-time scaling is particularly appealing in domains where verification remains easier than generation, making integer factorization a good match for the overall approach.

Given that the optimal test time compute allocation strategy depends, in principle, on the difficulty distribution across samples, uniform allocation is highly ineffective when that distribution is fairly mixed. In our case, however, we presume that we are solving a problem at a particular (fixed) bit-length, and since the factors themselves are sampled to be large, the intrinsic factorization difficulty remains fairly uniform, at each specific bit-length. We are therefore not concerned with adaptively allocating compute per-example, rather aiming to understand how much test-time compute is required in expectation to successfully factor numbers at increasing bit-lengths, for varying quality of the underlying model, expressed in its denoising negative log-likelihood. We evaluate both axes of test-time compute scaling empirically: scaling the denoising step budget of single-trajectory ancestral diffusion on long-horizon checkpoints in Section~\ref{sec:inference_scaling} (Table~\ref{tab:factorization_accuracy} and Figures~\ref{fig:scaling_law_denoise_steps}--\ref{fig:scaling_law_solve_rate_vs_loss}), and scaling the beam search frontier width with oracle self-conditioning across pre-training stages in Appendix~\ref{app:beam_search_scaling} (Figures~\ref{fig:bfs_frontier_by_stage}--\ref{fig:bfs_flops}).

\section{Results}
\label{sec:results}

Here we present a variety of empirical scaling laws for semiprime factorization using discrete diffusion models along the dimensions of data, model size, problem size, and test time compute. Additional plots, including those measured in terms of train time compute, are shown in the appendix.

\subsection{Data}

In Figure \ref{fig:scaling_law_data}, we plot validation loss against training steps (which is equivalent to data seen, since all models use the same batch size).\footnote{A batch size of $2^{14} = 16{,}384$ was used; hence for small factor bit-lengths ($k \le 19$), training involves multiple passes over the training set, see Section~\ref{sec:method}.}
We see that the loss curves exhibit power law behavior.
Unlike natural language, since prime factorization is a noiseless problem, it is no surprise that the y-intercepts of the fitted power laws can and often are 0. Importantly, because validation NLL averages over all $2k$ factor positions (both masked and unmasked) across noise levels $\eta \sim 2\eta$ with importance weights $w_b = 1/(2\eta_b)$, a \emph{non-factoring local-arithmetic baseline}—which simply copies visible unmasked bits and propagates revealed MSB/LSB prefixes and suffixes across mismatched mask offsets without global factoring (derived in Appendix~\ref{app:non_factoring_baseline})—already achieves an importance-weighted validation $\mathrm{NLL} \approx 0.262$. Consequently, validation NLL values above $\approx 0.262$ reflect learning token copying and local carry/division consistency on partially unmasked canvases, whereas genuine $\eta \to 1$ factorization emerges only once validation NLL drops below $\approx 0.262$.

\begin{figure}[htbp]
    \centering
    \includegraphics[width=\linewidth]{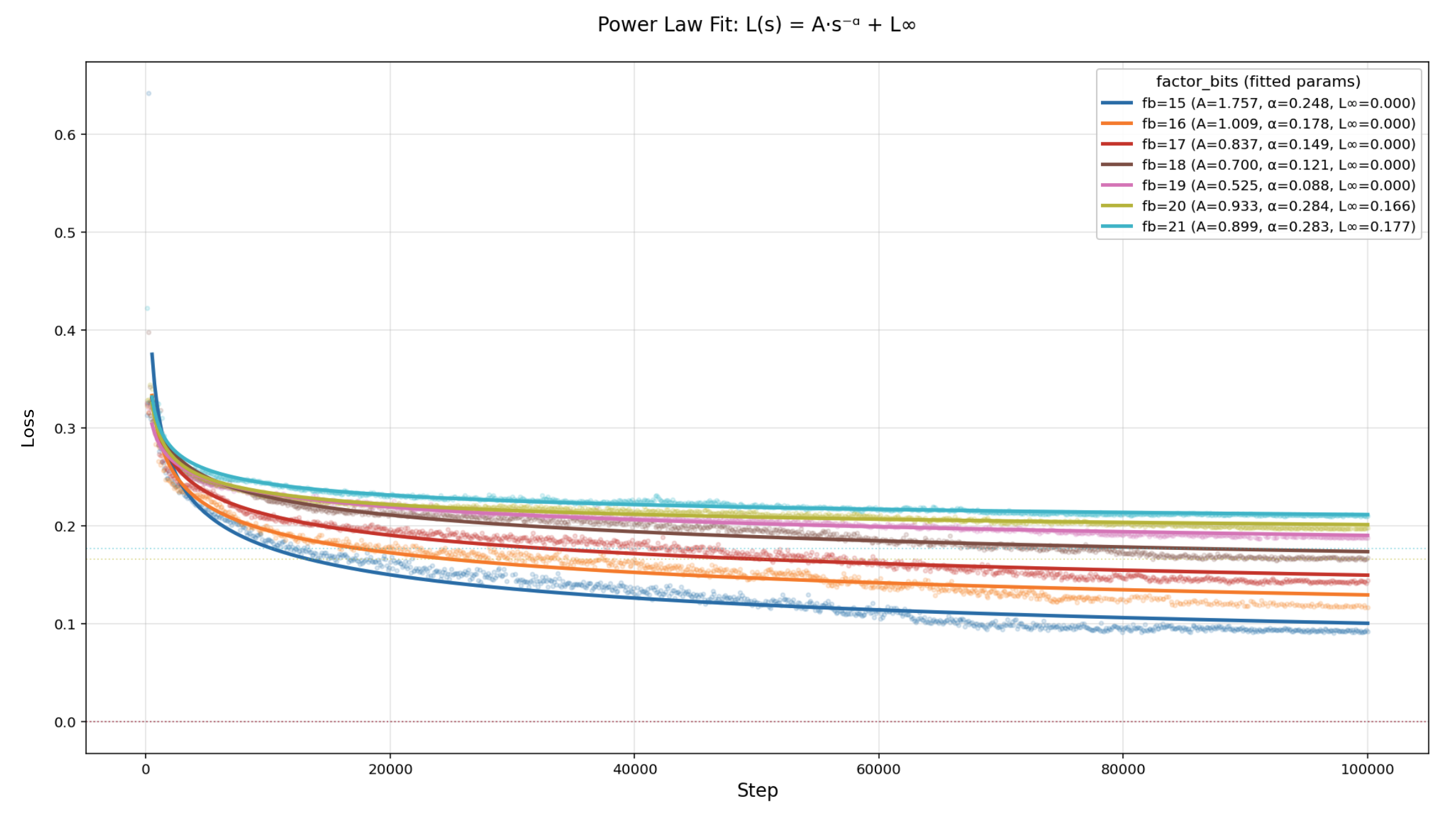}
    \caption{Validation negative log-likelihood (NLL) across training steps $s$ (with power-law fits $L(s) = A s^{-\alpha} + L_\infty$) for the 325M model across factor bit-lengths $k \in \{15, \dots, 21\}$.}
    \label{fig:scaling_law_data}
\end{figure}

\subsection{Model Size}

\begin{figure}[htbp]
    \centering
    \includegraphics[width=\linewidth]{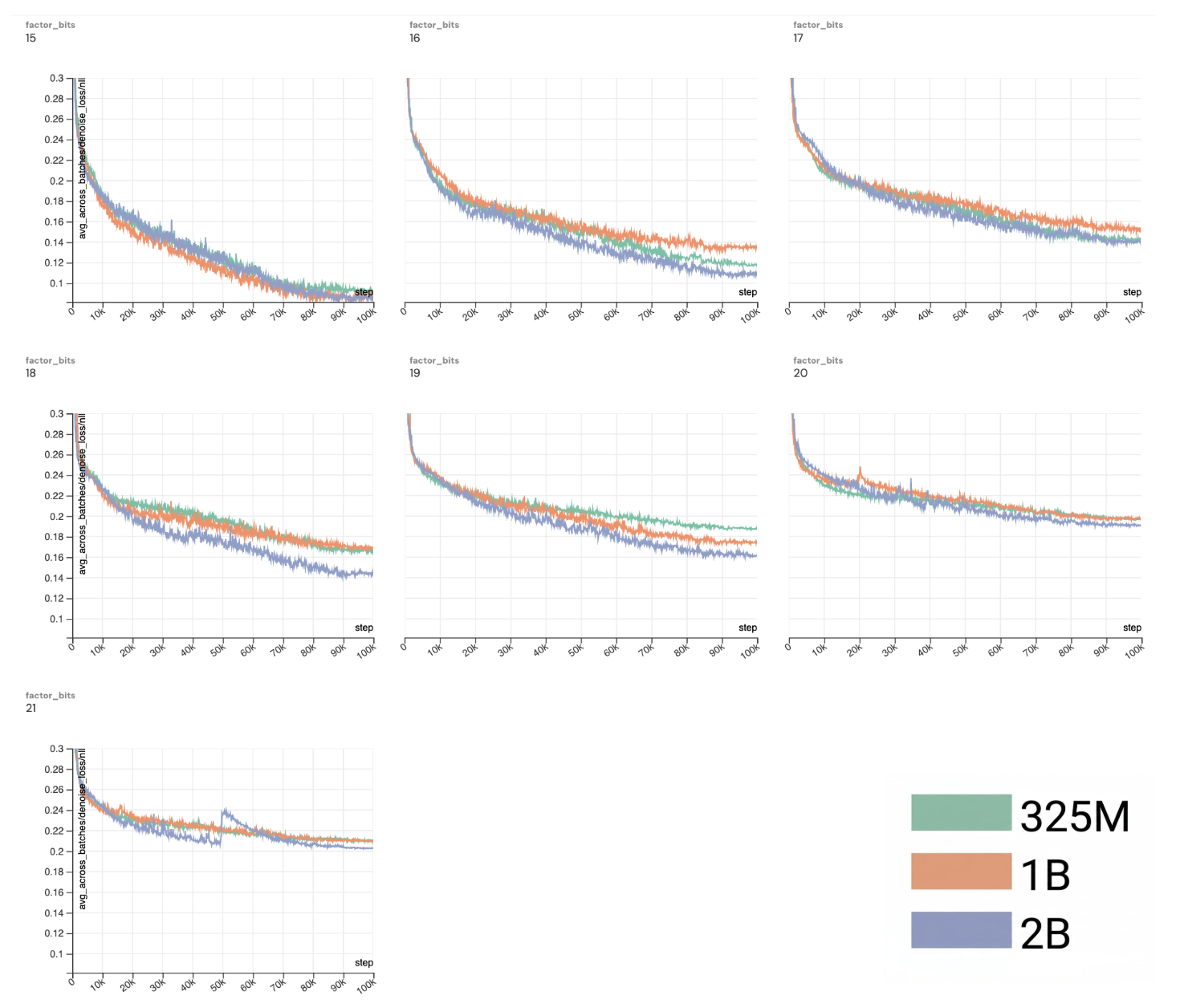}
    \caption{Validation NLL across training steps for three model sizes (325M, 1B, and 2B), faceted by factor bit-length $k \in \{15, \dots, 21\}$.}
    \label{fig:scaling_law_model_size}
\end{figure}

In Figure \ref{fig:scaling_law_model_size}, we plot validation loss curves against train steps for three model sizes (325M, 1B, 2B) across a range of factor bits. The decrease in the final loss value with model size is a bit noisy: while 2B is always (tied) for the best, the relationship between 325M and 1B is inconsistent, where which one has the lowest loss changes across factor bits.

\subsection{Inference-Time Scaling}
\label{sec:inference_scaling}

Here, we analyze how the performance of standalone single-trajectory ancestral diffusion (Section~\ref{sec:ancestral_sampling_main}; temperature $\tau = 0.7$, no beam search, oracle self-conditioning, or Coppersmith post-processing) scales with the number of inference-time denoising steps $T$ (beam search frontier scaling is analyzed separately in Appendix~\ref{app:beam_search_scaling}, Figures~\ref{fig:bfs_frontier_by_stage}--\ref{fig:bfs_flops}). As detailed in Section~\ref{sec:ancestral_sampling_main} and Appendix~\ref{app:ancestral_sampling}, the ancestral sampler performs no intermediate arithmetic check ($\hat{p} \cdot \hat{q} = n$) during the $T$ steps; instead, scaling $T \gg 2k$ improves solve rate via non-monotonic blocked-Gibbs refinement as fresh contiguous blocks of width $M_t = \max(\lfloor \eta_{t+1} k \rfloor, 1)$ are repeatedly re-masked and re-evaluated against the evolving soft self-conditioning state.

Table~\ref{tab:factorization_accuracy} reports single-trajectory exact factorization accuracy on the held-out test set ($1{,}024$ semiprimes) across $T \in [1024, 16384]$ for three trained checkpoints: a 325M model trained for $100{,}000$ steps at $k=15$, a 325M model trained for $347{,}800$ steps at $k=20$, and a 12B-parameter model trained for $109{,}000$ steps at $k=34$. Whenever a model factorizes a non-zero fraction of held-out semiprimes at $T = 1024$, accuracy improves roughly log-linearly with $\log T$; conversely, models with zero accuracy at low budgets ($k=34$) show no exact-factorization gains up to $T = 16{,}384$. For comparison at $k=15$, trial division over odd candidates up to $2^{15}$ guarantees success in at most $2^{14} = 16{,}384$ divisions.

\begin{table}[htbp]
  \centering
  \caption{Single-trajectory ancestral diffusion exact factorization accuracy on the held-out test set ($1{,}024$ test instances per bit-length; no beam search or Coppersmith post-processing; 1 sample per test instance, $\tau = 0.7$) across denoising step budgets $T$. Models: 325M trained for $100\text{k}$ steps at $k=15$, 325M trained for $348\text{k}$ steps at $k=20$, and 12B trained for $109\text{k}$ steps at $k=34$.}
  \label{tab:factorization_accuracy}
  \begin{tabular}{cccccc}
    \toprule
    \multirow{2}{*}{\textbf{Factor Bits}} & \multicolumn{5}{c}{\textbf{Diffusion Steps}} \\
    \cmidrule(lr){2-6}
    & \textbf{1024} & \textbf{2048} & \textbf{4096} & \textbf{8192} & \textbf{16384} \\
    \midrule
    15 & 64.55\% & 68.85\% & 76.47\% & 80.18\% & 87.60\% \\
    20 & 1.86\% & 3.42\% & 4.59\% & 6.35\% & 8.30\% \\
    34 & 0.00\% & 0.00\% & 0.00\% & 0.00\% & 0.00\% \\
    \bottomrule
  \end{tabular}
\end{table}

\begin{figure}[htbp!]
    \centering
    \includegraphics[width=\linewidth]{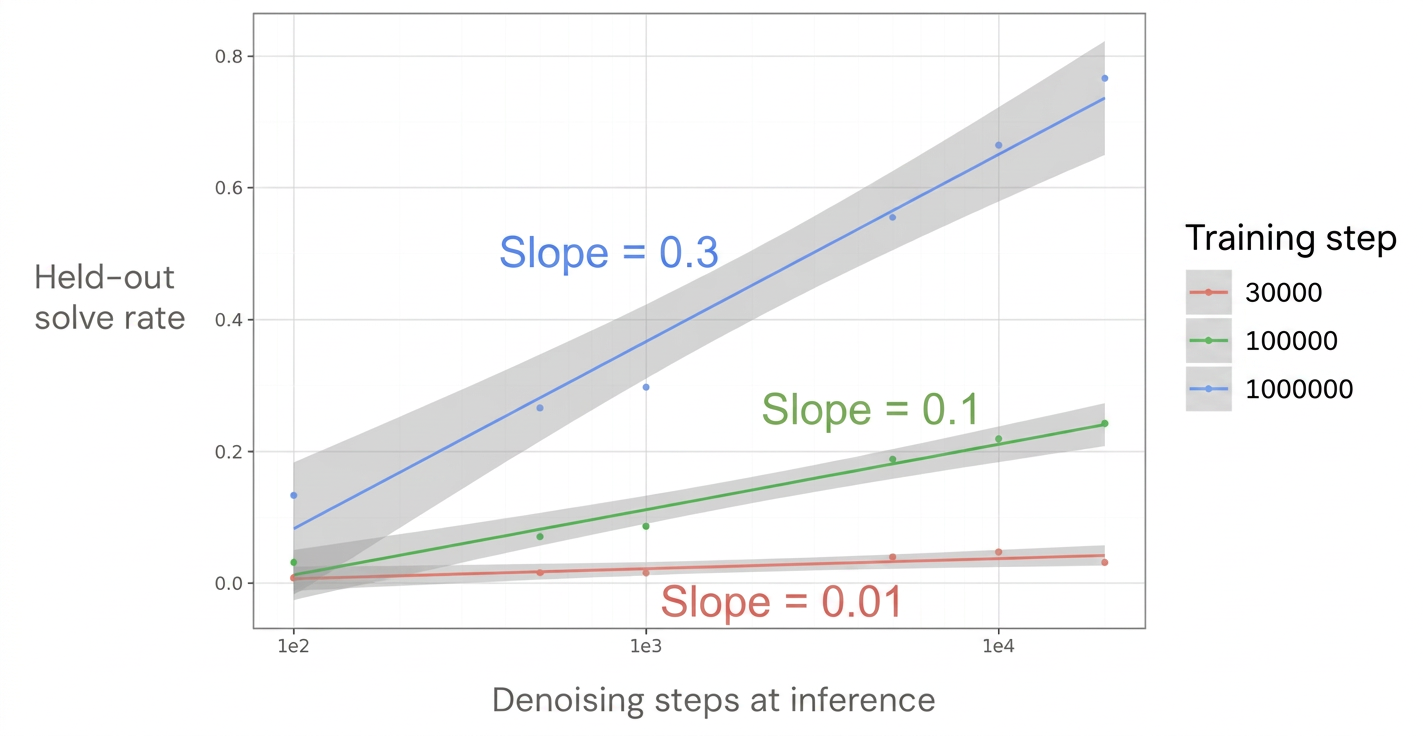}
    \caption{Log-linear scaling of single-trajectory ancestral solve rate with inference denoising steps $T$ across training horizons ($30\text{k}$, $100\text{k}$, $1\text{M}$ steps) for a 325M model at $k=15$ trained with our initial pointwise-corruption baseline recipe (prior to contiguous blockwise masking and two-pass concave rollout used in Figure~\ref{fig:scaling_law_solve_rate_vs_loss} and Table~\ref{tab:factorization_accuracy}).}
    \label{fig:scaling_law_denoise_steps}
\end{figure}

\begin{figure}[htbp]
    \centering
    \includegraphics[width=\linewidth]{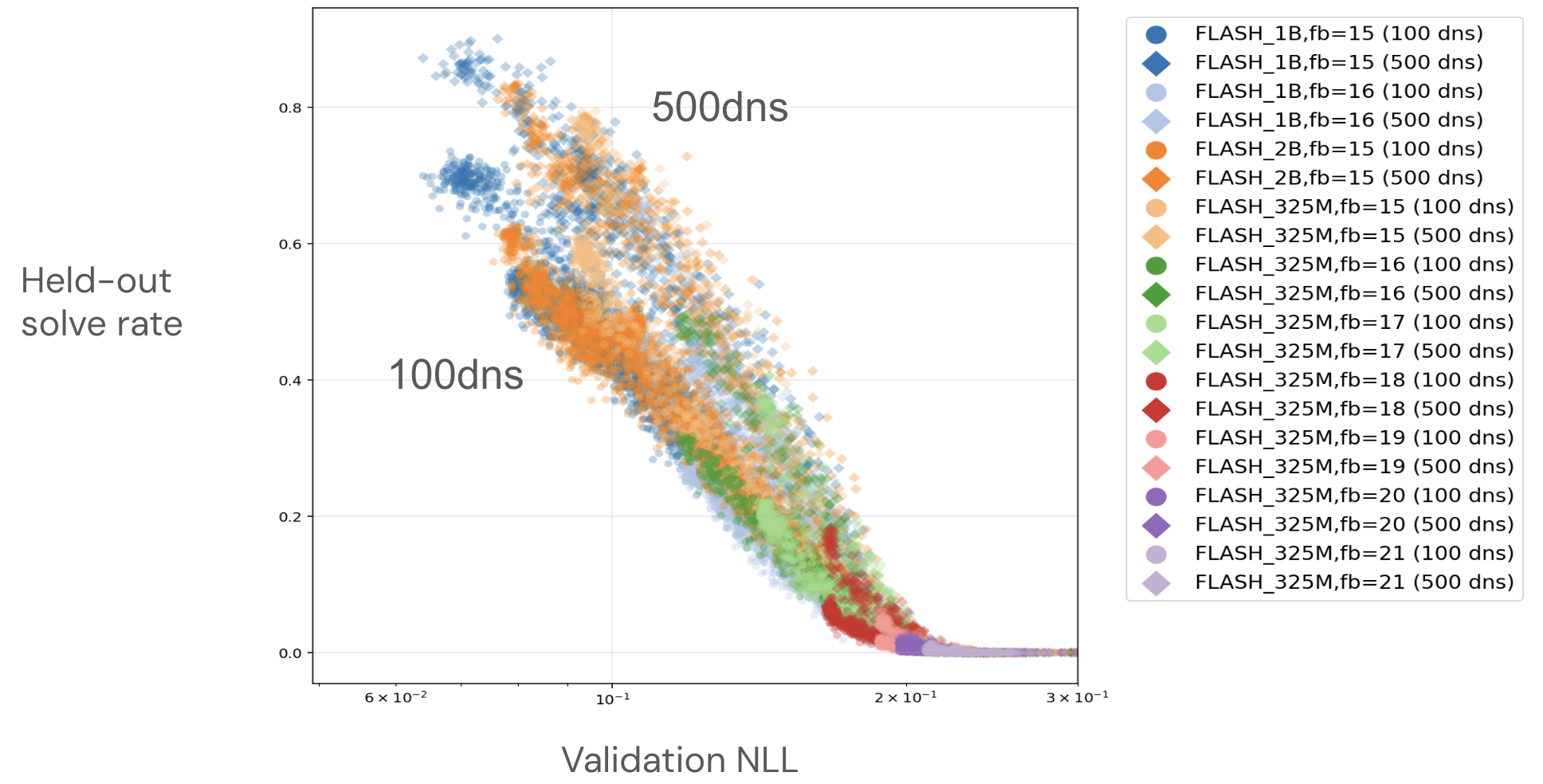}
    \caption{Each point on the above plot is a model's solve rate and validation loss at a particular point in training. The lower the validation loss, the higher the solve rate for any fixed number of denoising steps. This is true across various model sizes.}
    \label{fig:scaling_law_solve_rate_vs_loss}
\end{figure}

In Figure~\ref{fig:scaling_law_denoise_steps}, we illustrate this log-linear relationship across training horizons ($30\text{k}$, $100\text{k}$, and $1\text{M}$ steps) on an initial $k=15$ (325M) baseline trained with pointwise token corruption prior to introducing contiguous blockwise masking and two-pass concave rollout resampling; better-trained checkpoints exhibit steeper test-time scaling slopes. Figure~\ref{fig:scaling_law_solve_rate_vs_loss} evaluates checkpoints trained with our final blockwise-masking and two-pass rollout recipe (reaching $\approx 0.75$ accuracy at $T=100$ steps within $100\text{k}$ training steps for $k=15$) and demonstrates across model sizes and bit-lengths that single-trajectory solve rate at fixed $T \in \{100, 500\}$ is tightly governed by validation NLL. Notably, solve rate in Figure~\ref{fig:scaling_law_solve_rate_vs_loss} remains identically zero for all checkpoints with validation $\mathrm{NLL} \gtrsim 0.262$ and rises sharply only after NLL penetrates below $\approx 0.262$—matching the exact theoretical NLL floor of our non-factoring local-arithmetic baseline (Appendix~\ref{app:non_factoring_baseline}).

\subsection{Problem Size}

\begin{figure}[htbp]
    \centering
    \includegraphics[width=\linewidth]{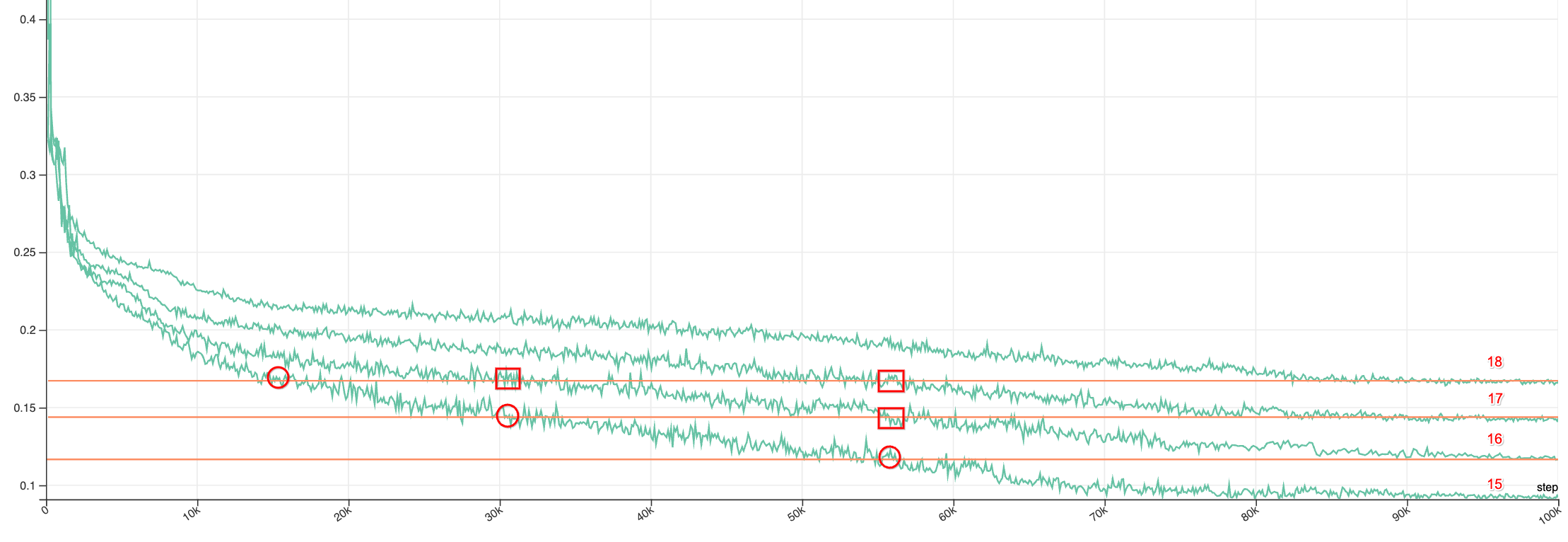}
    \caption{Loss curves for Flash 325M trained on factor bits 15, 16, 17, 18. We look at the final loss values of factor bits 16, 17, 18 and trace horizontal lines showcasing at what step loss curves corresponding to lower factor bits achieve those loss levels. Circled are approximate points along the loss curve for factor bit 15 where the horizontal lines meet it, corresponding to points in training where it achieves the final loss that the factor bits 16, 17, 18 problem achieves. Square boxes denote analogous approximate points of intersection among the other loss curves.}
    \label{fig:scaling_law_problem_size}
\end{figure}

Our prime factorization setup gives us another dimension for which to analyze scaling: problem size (factor bits).  Despite favorable scaling laws for data and test-time compute, we encounter a hurdle with problem size. In Figure \ref{fig:scaling_law_problem_size}, we plot loss curves for Flash 325M trained on factor bits 15, 16, 17, 18.  What we see is roughly a geometric relation between factor bits and a given loss value below the $\approx 0.262$ non-factoring local-arithmetic threshold (Appendix~\ref{app:non_factoring_baseline}); each increase in factor bit requires roughly a 2x increase in training steps to achieve a given loss value.
This trend suggests that data requirements scale exponentially in factor bits, i.e. linearly in the size of the problem. If this trend holds, this suggests that our diffusion model is unable to exploit  substructure from a sparse amount of data, and requires seeing a large fraction of semiprime examples in order to perform well. In essence, our modeling component scales similarly to brute force methods.

\subsection{Evaluation at 68 Bits ($k=34$) with Coppersmith Post-Processing}
\label{sec:factorization_68bit}

To probe the upper boundary of our training recipe, we trained a 12B-parameter diffusion model at $k = 34$ factor bits for $109$K steps at batch size $2^{14} = 16{,}384$ ($1.79 \times 10^9$ training examples, a negligible fraction of the $\approx 2.9 \times 10^{17}$ distinct 34-bit-bounded semiprimes). As shown in Table~\ref{tab:factorization_accuracy}, standalone ancestral sampling achieves $0\%$ factorization at $k = 34$ even with $T = 16{,}384$ denoising steps. However, Coppersmith's lattice-reduction method (Section~\ref{sec:post_processing}) requires a candidate factor to match only the leading $\lfloor k/2 \rfloor + 1 = 18$ MSBs to recover the remaining $16$ bits in polynomial time.

For this 68-bit experiment, we evaluated across a $100{,}000$-example held-out test split using single-trajectory ancestral diffusion ($T = 500$ denoising steps, $\tau = 0.7$, without beam search), generating \emph{one} candidate pair $(\hat{p}, \hat{q})$ per test instance and running Coppersmith reduction ($18$-bit prefix) on $\hat{p}$ and $\hat{q}$ ($2$ factor candidates tested per instance). Because our data generator samples primes $< 2^{34}$ without forcing the $34$-th bit to $1$ (Section~\ref{sec:method}), $25{,}228$ of the $100{,}000$ held-out instances have two full 34-bit prime factors ($p, q \in [2^{33}, 2^{34})$), and $15{,}523$ instances are strict 68-bit semiprimes ($n \in [2^{67}, 2^{68})$).

Out of the $15{,}523$ strictly 68-bit held-out semiprimes, Coppersmith post-processing succeeded on $1$ instance ($0.0064\%$; $1 / 25{,}228 = 0.0040\%$ across all pairs with two 34-bit factors), factoring
\[
182990651592147505079 = 15850613929 \times 11544704351,
\]
where $\hat{q}$ matched the leading $19 \ge 18$ MSBs of $q = 11544704351$ (and $\hat{p}$ matched $8$ MSBs of $p$).

\begin{table}[htbp]
\centering
\caption{Empirical leading-MSB prefix match distribution $P(L_{\mathrm{match}} \ge m)$ for the 12B model at $k=34$ ($1$ ancestral sample $(\hat{p}, \hat{q})$ per instance, $T=500$, across $N = 15{,}523$ held-out 68-bit semiprimes with $n \ge 2^{67}$), compared against (A) naive uniform random bit-guessing on $2$ factor candidates with fixed leading bit $1$ ($1 - (1 - 2^{-(m-1)})^2$), and (B) a zero-learning hyperbola magnitude prior that samples a random point on the product hyperbola $\hat{p}\hat{q} \approx n$ with $\hat{p} \ge \hat{q}$ by drawing $\hat{p}$ uniformly from the valid integer interval $\{\lceil\sqrt{n}\rceil, \dots, 2^{34}-1\}$ and setting $\hat{q} = \lfloor n / \hat{p} \rfloor$.}
\label{tab:msb_distribution_34bit}
\small
\setlength{\tabcolsep}{4.5pt}
\begin{tabular}{r r r r r}
\toprule
\textbf{MSBs ($m$)} & \textbf{12B Count} & \textbf{12B Rate} & \textbf{Uniform Prior} & \textbf{Hyperbola Prior} \\
\midrule
$\ge 2$  & $15{,}489$ & $99.78\%$  & $75.00\%$  ($11{,}642.3$) & $100.00\%$ ($15{,}523.0$) \\
$\ge 4$  & $7{,}957$  & $51.26\%$  & $23.44\%$  ($3{,}638.2$)  & $47.84\%$  ($7{,}426.9$)  \\
$\ge 6$  & $2{,}526$  & $16.27\%$  & $6.15\%$   ($955.0$)      & $14.74\%$  ($2{,}288.4$)  \\
$\ge 8$  & $701$      & $4.516\%$  & $1.556\%$  ($241.6$)      & $3.924\%$  ($609.2$)      \\
$\ge 10$ & $189$      & $1.218\%$  & $0.390\%$  ($60.6$)       & $0.978\%$  ($151.8$)      \\
$\ge 12$ & $49$       & $0.316\%$  & $0.098\%$  ($15.2$)       & $0.238\%$  ($37.0$)       \\
$\ge 14$ & $14$       & $0.090\%$  & $0.024\%$  ($3.8$)        & $0.059\%$  ($9.1$)        \\
$\ge 16$ & $5$        & $0.032\%$  & $0.0061\%$ ($0.95$)       & $0.0142\%$ ($2.20$)       \\
$\ge 18$ & $1$        & $0.0064\%$ & $0.0015\%$ ($0.24$)       & $0.0040\%$ ($0.62$)       \\
$\ge 19$ & $1$        & $0.0064\%$ & $0.0008\%$ ($0.12$)       & $0.0020\%$ ($0.31$)       \\
\bottomrule
\end{tabular}
\end{table}

To assess statistical significance against random prefix coincidence, Table~\ref{tab:msb_distribution_34bit} compares the model's empirical MSB match distribution across all $15{,}523$ full 68-bit instances against two non-factoring baselines: (A) naive independent uniform bit-guessing in $[2^{33}, 2^{34})$, and (B) a zero-learning \emph{hyperbola magnitude prior}. Because any valid ordered factor pair $p \ge q$ of $n = p \cdot q$ with $p, q < 2^{34}$ must satisfy $p \in [\lceil\sqrt{n}\rceil, 2^{34}-1]$ and lie on the rectangular hyperbola $q = n/p$, the hyperbola prior simply draws $\hat{p}$ uniformly at random from the valid integer interval $\{\lceil\sqrt{n}\rceil, \dots, 2^{34}-1\}$ and sets $\hat{q} = \lfloor n / \hat{p} \rfloor$ by division. Across intermediate prefix lengths ($m \in [4, 16]$), the 12B denoiser consistently matches leading prefixes at $\approx 3\times$ the rate of naive uniform bit-guessing (e.g., $189$ vs.\ $60.6$ expected matches at $m \ge 10$, $z > 16$). Comparing against the hyperbola prior reveals the source of this lift: roughly $2.5\times$ arises purely from restricting $(\hat{p}, \hat{q})$ to the product hyperbola $\hat{p} \cdot \hat{q} \approx n$ with $\hat{p} \ge \hat{q}$, with a modest additional $\approx 1.25\times$ concentration over uniform hyperbola sampling ($189$ vs.\ $151.8$ at $m \ge 10$). Beyond bit $m = 4$, however, each additional bit halves the empirical survival rate ($\approx 0.52\times$ per bit), mirroring random guessing on deeper factor bits. Consequently, at the Coppersmith threshold $m = 18$, a single observed hit across $15{,}523$ single-sample trials (compared to $0.24$ expected under naive uniform guessing and $0.62$ under the hyperbola prior) reflects tail sampling from a magnitude-calibrated prior rather than reliable 68-bit algorithmic factorization—consistent with our exponential scaling law.

\section{Conclusion}
\label{sec:conclusion}

Through our experiments, we have identified several key insights. At small bit lengths ($k \le 20$), our models are able to reach high semiprime factoring performance. Our training setup rules out direct memorization, leading us to conclude that the denoiser can learn to successfully solve the task given sufficient training data. Due to the overlap in the constituent prime factor pool that is used to derive semiprimes for training and testing, the model benefits from having encountered the same factors previously in different numbers. The size of this prime factor vocabulary scales as $\pi(2^k) \approx 2^k / (k \ln 2)$. This scaling aligns with our empirical finding that achieving a target validation loss requires a multiplicative growth in training data and compute ($\approx 2\times$) for each added factor bit. Our scaling results also indicate that the model fails to uncover any polynomial or sub-exponential factoring algorithm, scaling approximately as a kind of learned brute force search.

It is instructive to contrast this performance with the scaling dynamics arising in natural language modeling, where smooth power-law scaling and length generalization occur. This generalization stems from a rich, hierarchical, and locally constrained structure that the model is able to learn and exploit. In contrast, semiprime factorization is globally coupled and cryptographically pseudorandom; a single bit-flip in the semiprime number has discontinuous, chaotic effects on its factors. The lack of reusable local substructure makes it virtually impossible for a diffusion model to generalize from $k$-bit instances to $k+1$-bit instances without a prohibitive, exponential increase in data.

Finally, our 68-bit MSB distribution analysis reveals that while the denoiser reliably learns coarse multiplicative magnitude bounds ($\approx 3\times$ prefix-match lift over naive random bits), its accuracy on deeper bits decays at near-chance rates. We therefore conclude that direct bit-prediction ML methods pose no present or future threat to RSA cryptography.

\bibliography{main}

\appendix

\section{Architectural, Training, and Test-Time Search Details}
\label{app:architecture_and_search}

In this appendix, we detail the input representations,
self-conditioning architecture, importance-weighted diffusion training objective,
and test-time beam search with oracle self-conditioning used in our experiments. We are unable to provide full architectural details, since we used proprietary Flash models as our transformer-based diffusion model.

\subsection{Canvas Representation}
\label{app:positional_encodings}

Each factorization instance consists of a $2k$-bit semiprime $n = p \cdot q$ and
its two $k$-bit prime factors $p \ge q$, serialized into a single sequence (or
\emph{canvas}) of length $L = 4k$:
\begin{equation}
  \mathbf{x} = \big(\underbrace{x_1, \dots, x_{2k}}_{n\text{ ($2k$ bits)}}, \;
  \underbrace{x_{2k+1}, \dots, x_{3k}}_{p\text{ ($k$ bits)}}, \;
  \underbrace{x_{3k+1}, \dots, x_{4k}}_{q\text{ ($k$ bits)}}\big) \in \mathcal{V}^L,
\end{equation}
where $\mathcal{V}$ denotes the discrete token vocabulary containing binary bit
tokens $\{0, 1\}$, the absorbing mask token, and the pad token ($|\mathcal{V}| = 4$). Let $L_{\text{off}} = 2k$ denote the
prefix offset corresponding to the conditioned semiprime $n$.

\paragraph{Token Embedding Normalization.}
Let $\mathbf{E} \in \mathbb{R}^{|\mathcal{V}| \times d}$ denote the learnable
token embedding table with model dimension $d$. Each row of $\mathbf{E}$ is
projected onto the unit Euclidean ($\ell_2$) sphere ($\|\hat{\mathbf{E}}_v\|_2 = 1$,
corresponding to root-mean-square $1/\sqrt{d}$):
\begin{equation}
  \hat{\mathbf{E}}_v = \frac{\mathbf{E}_v}{\sqrt{\sum_{j=1}^d E_{v,j}^2 + \epsilon}},
  \qquad \epsilon = 10^{-30}.
  \label{eq:token_emb_norm}
\end{equation}
For a discrete noisy canvas $\tilde{\mathbf{x}} \in \mathcal{V}^L$, the raw
token embeddings are scaled by $\sqrt{d}$ upon lookup so that each token
embedding has Euclidean norm $\sqrt{d}$ (unit root-mean-square):
\begin{equation}
  \mathbf{C} = \mathbf{H}_{\text{tok}}(\tilde{\mathbf{x}}) = \sqrt{d}\,\mathbf{O}(\tilde{\mathbf{x}})\hat{\mathbf{E}} \in \mathbb{R}^{L \times d},
  \label{eq:canvas_emb}
\end{equation}
where $\mathbf{O}(\tilde{\mathbf{x}}) \in \{0,1\}^{L \times |\mathcal{V}|}$ is
the one-hot encoding of $\tilde{\mathbf{x}}$.

\subsection{Self-Conditioning Architecture}
\label{app:self_conditioning}

To allow the discrete diffusion denoiser to refine its factor estimates across
iterations without losing continuous uncertainty information when discretizing to
hard tokens, we incorporate a gated feed-forward self-conditioning module.

\paragraph{Soft-Logit Embedding.}
Given unnormalized vocabulary logits $\boldsymbol{\ell} \in \mathbb{R}^{L \times |\mathcal{V}|}$
produced by a preceding denoising pass, we compute a continuous probability-weighted
embedding over the shared normalized token embedding table $\hat{\mathbf{E}}$:
\begin{equation}
  \mathbf{S}(\boldsymbol{\ell}) = \sqrt{d} \, \mathrm{softmax}(\boldsymbol{\ell}) \hat{\mathbf{E}}
  \in \mathbb{R}^{L \times d}.
  \label{eq:embed_logits}
\end{equation}
Given that $\mathbf{H}_{\text{tok}}$ and $\mathbf{S}$ project onto the
exact same normalized basis $\sqrt{d}\,\hat{\mathbf{E}}$, a degenerate (one-hot)
probability distribution in Eq.~\eqref{eq:embed_logits} recovers the exact
discrete token embedding in $\mathbf{H}_{\text{tok}}$.

\paragraph{Prefix Zeroing and Gated FFN Transformation.}
Because the first $L_{\text{off}} = 2k$ positions correspond to the known
semiprime $n$, the self-conditioning module first masks out the prefix positions:
\begin{equation}
  \tilde{\mathbf{S}}_i =
  \begin{cases}
    \mathbf{0} & \text{if } 1 \le i \le 2k, \\
    \mathbf{S}_i & \text{if } 2k < i \le 4k.
  \end{cases}
\end{equation}
Rather than concatenating $\tilde{\mathbf{S}}$ with the canvas representations
along the channel dimension, $\tilde{\mathbf{S}}$ is transformed by a dedicated
pre-normalized gated GeLU feed-forward network ($\mathrm{FFWSubBlock}$) with
hidden expansion dimension $d_{\text{ffw}} = 4d$ and zero bias:
\begin{align}
  \mathbf{U} &= \mathrm{RMSNorm}_{\gamma}(\tilde{\mathbf{S}}), \\
  \mathbf{G}_{\text{SC}}(\tilde{\mathbf{S}}) &=
  \Big(\mathrm{GeLU}(\mathbf{U}\mathbf{W}_A) \odot (\mathbf{U}\mathbf{W}_B)\Big)\mathbf{W}_C,
  \label{eq:sc_ffn}
\end{align}
where $\mathbf{W}_A, \mathbf{W}_B \in \mathbb{R}^{d \times 4d}$ and
$\mathbf{W}_C \in \mathbb{R}^{4d \times d}$ are learnable projection matrices
and $\odot$ denotes elementwise multiplication. Because all linear projections
in Eq.~\eqref{eq:sc_ffn} are bias-free and $\mathrm{GeLU}(0) = 0$, any zero
vector in $\tilde{\mathbf{S}}$ (such as the $2k$-bit prefix $n$, or the entire
self-conditioning tensor on the first denoising step) maps identically to
$\mathbf{0} \in \mathbb{R}^d$.

\paragraph{Residual Combination and Output Normalization.}
The transformed self-conditioning signal $\mathbf{G}_{\text{SC}}(\tilde{\mathbf{S}})$
is combined with the pre-normalized canvas embeddings $\mathbf{C}$ from
Eq.~\eqref{eq:canvas_emb} via elementwise residual addition, followed by a
scale-free $\mathrm{RMSNorm}$ layer before entering the $N_{\text{layer}}$-block
bidirectional transformer trunk:
\begin{equation}
  \mathbf{Z}^{(0)} = \mathrm{RMSNorm}\big(\mathbf{C} + \mathbf{G}_{\text{SC}}(\tilde{\mathbf{S}})\big).
  \label{eq:sc_combine}
\end{equation}
After passing $\mathbf{Z}^{(0)}$ through the bidirectional transformer layers,
the final hidden states $\mathbf{Z}^{(N_{\text{layer}})} \in \mathbb{R}^{L \times d}$
are unembedded via transposed projection against $\hat{\mathbf{E}}^\top$ with
$\tanh$ logit soft-capping ($c = 30.0$) to yield the output logits. Note that
the transformer denoiser takes no explicit scalar timestep or noise-level
embedding $\eta_t$; instead, it infers the current progress of decoding directly
from the spatial pattern of mask tokens on $\tilde{\mathbf{x}}$ and the sharpness
of the continuous self-conditioning embeddings $\tilde{\mathbf{S}}$.

\subsection{Two-Pass Training Rollout and Noise-Level Loss Weighting}
\label{app:training_rollout_and_loss}

\paragraph{Importance-Weighted Noise Sampling and Blockwise Masking.}
During training, each example $b \in \{1, \dots, B_{\text{batch}}\}$ in a minibatch is assigned
a noise proportion $\eta_b \in [\eta_{\min}, 1]$ with $\eta_{\min} = 0.02$. To
reduce gradient variance without biasing the diffusion objective, we employ
importance sampling with a triangular proposal distribution $p(\eta) \propto 2\eta$
on $[\eta_{\min}, 1]$. Specifically, we draw $u_b \sim \mathcal{U}(\eta_{\min}^2, 1)$
and set $\eta_b = \sqrt{u_b}$ via inverse-transform sampling, pairing each
example with its inverse-density importance weight:
\begin{equation}
  \eta_b = \sqrt{u_b}, \quad u_b \sim \mathcal{U}(\eta_{\min}^2, 1),
  \qquad w_b = \frac{1}{2\eta_b}.
  \label{eq:importance_weights}
\end{equation}
Given $\eta_b$, the forward masking process leaves the semiprime prefix
$x_{b, 1:2k}$ intact and masks two independently positioned contiguous blocks of deterministic length
$M_b = \max(\lfloor \eta_b k \rfloor, 1)$
starting at uniformly sampled bit offsets $s_{b,p}, s_{b,q} \sim \mathcal{U}\{0, \dots, k - M_b\}$
within $p$ and $q$, respectively, yielding the corrupted canvas $\tilde{\mathbf{x}}_b$.

\paragraph{Two-Pass Self-Conditioning Rollout with Concave Canvas Resampling.}
Standard self-conditioning~\citep{chen2023analogbitsgeneratingdiscrete} randomly
chooses with $50\%$ probability whether to run a single unconditioned pass
($\mathbf{S}^{(0)} = \mathbf{0}$) or a two-pass self-conditioned rollout. In our
pipeline, we instead execute a two-pass rollout on \emph{every} training step so
that Pass~2 can simultaneously train self-conditioning and expose the denoiser to
partially resampled discrete canvases:
\begin{enumerate}
  \item \textbf{Pass 1 (Unconditioned Proposal):} With dropout disabled, the
  denoiser is evaluated on $\tilde{\mathbf{x}}_b$ with an all-zero
  self-conditioning tensor $\mathbf{S}^{(0)} = \mathbf{0}$, producing first-pass
  logits $\boldsymbol{\ell}_b^{(1)}$. No parameter gradients are backpropagated
  through Pass~1.
  \item \textbf{Soft Self-Conditioning and Concave Canvas Resampling:} We compute
  the soft self-conditioning embeddings
  $\mathbf{S}_b^{(1)} = \mathrm{stop\_gradient}\big(\mathbf{S}(\boldsymbol{\ell}_b^{(1)})\big)$
  via Eq.~\eqref{eq:embed_logits}. In addition, during iterative test-time
  inference and beam search, the denoiser must operate on canvases containing a
  mixture of masked tokens and partially committed discrete predictions (some of
  which may be erroneous guesses from earlier steps). To simulate this regime
  during training, we sample discrete tokens
  $\hat{\mathbf{x}}_b^{(1)} \sim \mathrm{Categorical}(\boldsymbol{\ell}_b^{(1)})$
  and replace each factor position $i > 2k$ of $\tilde{\mathbf{x}}_b$ (across
  \emph{both} masked positions and initially unmasked ground-truth factor
  positions) with $\hat{x}_{b,i}^{(1)}$ independently with a concave
  noise-dependent probability:
  \begin{equation}
    p_{\text{replace}}(\eta_b) = 4 \rho \, \eta_b (1 - \eta_b), \qquad \rho = 0.5,
    \label{eq:concave_replace}
  \end{equation}
  forming the resampled canvas $\tilde{\mathbf{x}}_b^{(2)}$. Note that
  $p_{\text{replace}}(\eta_b)$ vanishes at $\eta_b \in \{0, 1\}$ and peaks at
  $\rho = 0.5$ when $\eta_b = 0.5$. Overwriting both masked and unmasked factor
  bits teaches Pass~2 not only to inpaint mask tokens using $\mathbf{S}_b^{(1)}$,
  but also to detect and correct erroneous discrete bit commitments on the canvas.
  \item \textbf{Pass 2 (Self-Conditioned Update with $1\%$ Unconditioned Dropout):}
  Because parameter gradients are computed exclusively on Pass~2, always passing
  $\mathbf{S}_b^{(1)}$ would leave the denoiser untrained for the initial
  inference step $t = 0$, where no prior prediction exists ($\mathbf{S}^{(0)} = \mathbf{0}$).
  Therefore, on Pass~2, $\mathbf{S}_b^{(1)}$ is retained with Bernoulli
  probability $p_{\text{SC}} = 0.99$ and zeroed out ($\mathbf{S}_b^{(1)} \leftarrow \mathbf{0}$)
  with probability $1 - p_{\text{SC}} = 0.01$. The denoiser then predicts
  second-pass logits $\boldsymbol{\ell}_b^{(2)}$ from
  $(\tilde{\mathbf{x}}_b^{(2)}, \mathbf{S}_b^{(1)})$.
\end{enumerate}

\paragraph{Importance-Weighted Cross-Entropy Loss.}
Only Pass 2 contributes to the parameter gradient. The batch loss is computed as the
self-normalized importance-weighted cross-entropy:
\begin{equation}
  \mathcal{L}(\theta) =
  \frac{\sum_{b=1}^{B_{\text{batch}}} \sum_{i=2k+1}^{4k} w_b \,
  \ell_{\text{CE}}\big(x_{b,i}, \, \boldsymbol{\ell}_{b,i}^{(2)}(\theta)\big)}
  {2k \sum_{b=1}^{B_{\text{batch}}} w_b},
  \qquad w_b = \frac{1}{2\eta_b}.
  \label{eq:ce_loss}
\end{equation}

\subsubsection{Non-Factoring Local-Arithmetic Baseline}
\label{app:non_factoring_baseline}

Because Eq.~\eqref{eq:ce_loss} and validation token accuracy are averaged across \emph{all} $2k$ factor positions of $p$ and $q$—including both masked and unmasked positions—at noise levels $\eta_b \sim p(\eta) = 2\eta$ on $[0.02, 1]$, a model with \emph{zero global factoring ability} can still achieve a low validation NLL and high per-token accuracy. Specifically, consider an idealized \emph{non-factoring local-arithmetic baseline} operating on a partially masked canvas $(n, \tilde{p}, \tilde{q})$ that applies only three elementary local rules without attempting to factor $n$:
\begin{enumerate}
  \item \textbf{Copying visible unmasked factor bits ($\approx 33.3\%$ of all factor positions):}
  At noise level $\eta$, a contiguous block of width $M = \max(\lfloor \eta k \rfloor, 1)$ is masked in $p$ and in $q$, leaving the remaining $1 - M/k \approx 1 - \eta$ fraction of factor bits unmasked on $\tilde{\mathbf{x}}_b$. Under the triangular proposal $p(\eta) = 2\eta$ on $[0.02, 1]$, the expected masked fraction is
  \[
    \mathbb{E}[\eta] \approx \int_0^1 \eta \cdot (2\eta)\,d\eta = \frac{2}{3},
  \]
  meaning $1 - \mathbb{E}[\eta] \approx \frac{1}{3}$ ($33.3\%$) of the $2k$ factor bits are already visible on the input canvas. Even a pure token-copying baseline that copies the visible $\frac{1}{3}$ of bits ($\mathrm{NLL} = 0$, $\mathrm{Accuracy} = 1$) and predicts $50/50$ coin flips ($\mathrm{NLL} = \ln 2 \approx 0.693$, $\mathrm{Accuracy} = 0.5$) on the masked $\frac{2}{3}$ of bits already achieves unweighted $\mathrm{NLL} = \frac{2}{3}\ln 2 \approx 0.462$ and $\mathrm{Accuracy} = \frac{1}{3}(1) + \frac{2}{3}(0.5) = 66.7\%$.
  \item \textbf{Odd parity on the two LSBs:}
  Every prime $> 2$ is odd, so the least significant bits $p_{k-1} = 1$ and $q_{k-1} = 1$ are deterministically $1$ even when covered by the mask.
  \item \textbf{Cross-factor MSB division and LSB modular multiplication across mismatched mask offsets:}
  Because the contiguous mask starting offsets $s_{b,p}, s_{b,q} \sim \mathcal{U}\{0, \dots, k - M_b\}$ are drawn \emph{independently} in $p$ and $q$, the masked spans $[s_{b,p}, s_{b,p} + M_b)$ and $[s_{b,q}, s_{b,q} + M_b)$ rarely align:
  \begin{itemize}
    \item \emph{LSB suffix propagation ($p \cdot q \equiv n \pmod{2^r}$):} Whenever the mask in $q$ extends $r$ bits closer to the LSB than the mask in $p$ (or vice versa), the $r$ lowest bits of $p$ are unmasked while the corresponding $r$ lowest bits of $q$ are masked. Since odd primes are invertible modulo $2^r$, schoolbook multiplication and carry propagation uniquely determine those $r$ masked LSBs of $q \equiv n \cdot p^{-1} \pmod{2^r}$.
    \item \emph{MSB prefix propagation ($q = n / p$):} Symmetrically, whenever $s_{b,p} > s_{b,q}$, the leading $s_{b,p}$ MSBs of $p$ are visible while bits $[s_{b,q}, s_{b,p})$ of $q$ are masked; long division of $n$ by the visible prefix of $p$ deterministically resolves the corresponding masked MSBs of $q$ (up to boundary carry).
  \end{itemize}
  For two independent uniform draws $s_{b,p}, s_{b,q}$ on $[0, (1-\eta)k]$, the expected offset mismatch $\mathbb{E}[|s_{b,p} - s_{b,q}|]$ is $\frac{1}{3}(1-\eta)k$ bits on the MSB side and $\frac{1}{3}(1-\eta)k$ bits on the LSB side, allowing local cross-factor arithmetic to deterministically resolve an additional $\approx \frac{1}{3}(1 - \mathbb{E}[\eta]) \approx 11\text{--}13\%$ of all factor positions. On all remaining positions where both $p$ and $q$ are simultaneously masked beyond local carry reach, the baseline predicts a $50/50$ coin flip ($\mathrm{NLL} = \ln 2$, $\mathrm{Accuracy} = 0.5$).
\end{enumerate}

Integrating this non-factoring predictor over $\eta_b \sim 2\eta$ on $[0.02, 1]$ (including the Pass-2 concave token replacement $p_{\text{replace}}(\eta_b) = 2\eta_b(1-\eta_b)$ of Eq.~\ref{eq:concave_replace}) yields two exact reference benchmarks corresponding to the two validation weighting conventions used across our figures:
\begin{itemize}
  \item \textbf{Importance-Weighted Validation ($w_b = \frac{1}{2\eta_b}$, Main-Text Figures~\ref{fig:scaling_law_data}, \ref{fig:scaling_law_model_size}, \ref{fig:scaling_law_solve_rate_vs_loss}, and \ref{fig:scaling_law_problem_size}):}
  Multiplying each example by $w_b = \frac{1}{2\eta_b}$ cancels the $2\eta$ proposal density and heavily upweights low-noise examples ($\eta_b \to 0.02$, where nearly all factor bits are unmasked and locally recoverable), yielding:
  \begin{equation}
    \mathrm{NLL}_{\text{baseline}}^{\text{(weighted)}} \approx 0.262.
    \label{eq:nll_baseline_weighted}
  \end{equation}
  This explains the sharp phase transition in Figure~\ref{fig:scaling_law_solve_rate_vs_loss}: every checkpoint with weighted validation $\mathrm{NLL} \ge 0.262$ has $0.0\%$ exact solve rate, and non-zero factorization emerges only once validation NLL penetrates below $0.262$.
  \item \textbf{Unweighted Validation ($w_b = 1$, Appendix~\ref{sec:appendix_scaling} Figures~\ref{fig:isoflop}--\ref{fig:training_dynamics_325m_linear}):}
  Evaluating over $\eta_b \sim 2\eta$ on $[0.02, 1]$ with unit weights $w_b = 1$ yields:
  \begin{equation}
    \mathrm{NLL}_{\text{baseline}}^{\text{(unweighted)}} \approx 0.375,
    \qquad
    \mathrm{Accuracy}_{\text{baseline}}^{\text{(unweighted)}} \approx 0.732 \;\;(73.2\%).
    \label{eq:nll_baseline_unweighted}
  \end{equation}
  This explains why models at large bit-lengths ($k \ge 24$) in Figures~\ref{fig:isoflop}--\ref{fig:training_dynamics_325m_linear} plateau at $\mathrm{NLL} \approx 0.375\text{--}0.385$ and $\mathrm{Accuracy} \approx 73.2\%$: achieving $73.2\%$ validation bit accuracy reflects mastering token copying and local MSB/LSB carry propagation on partially unmasked canvases, while still guessing $50/50$ on the unfactored core.
\end{itemize}

\subsection{Test-Time Denoising, Beam Search, and Oracle Self-Conditioning}
\label{app:beam_search_oracle_sc}

Here we provide a full specification of our two test-time
decoding procedures: single-trajectory non-monotonic ancestral sampling and
guided MSB-to-LSB beam search with oracle self-conditioning.

\subsubsection{Non-Monotonic Ancestral Block-Diffusion Sampling}
\label{app:ancestral_sampling}

Given a $T$-step inference budget ($t = 0, 1, \dots, T-1$), ancestral sampling
initializes the canvas $\tilde{\mathbf{x}}^{(0)}$ with the $2k$ bits of $n$
unmasked and all $2k$ factor bits of $p$ and $q$ set to the mask token
($\eta_0 = 1$), alongside an all-zero self-conditioning tensor
$\mathbf{S}^{(0)} = \mathbf{0}$. At each step $t$, with incoming noise level
$\eta_t = \frac{T-t}{T}$ and target next-step noise level
$\eta_{t+1} = 1 - \frac{t+1}{T} = \frac{T-1-t}{T}$, the sampler executes four
operations:
\begin{enumerate}
  \item \textbf{Denoising Forward Pass:} The transformer denoiser evaluates
  $(\tilde{\mathbf{x}}^{(t)}, \mathbf{S}^{(t)})$ to produce vocabulary logits
  $\boldsymbol{\ell}^{(t)} \in \mathbb{R}^{4k \times |\mathcal{V}|}$.
  \item \textbf{Binary Logit Shaping and Temperature Scaling:} To prevent the
  sampler from proposing non-binary tokens (mask or pad) at factor positions,
  we form shaped logits $\tilde{\boldsymbol{\ell}}^{(t)}$ by setting the logits of
  all non-binary vocabulary entries to $-\infty$ and scaling by sampling
  temperature $\tau = 0.7$:
  \begin{equation}
    \tilde{\ell}_{j, v}^{(t)} =
    \begin{cases}
      \ell_{j, v}^{(t)} / \tau & \text{if } v \in \{0, 1\}, \\
      -\infty & \text{otherwise.}
    \end{cases}
  \end{equation}
  \item \textbf{Soft Self-Conditioning Update:} The continuous self-conditioning
  tensor for step $t+1$ is computed from the shaped binary logits via
  Eq.~\eqref{eq:embed_logits}:
  $\mathbf{S}^{(t+1)} = \mathbf{S}(\tilde{\boldsymbol{\ell}}^{(t)}) = \sqrt{d}\,\mathrm{softmax}(\tilde{\boldsymbol{\ell}}^{(t)})\hat{\mathbf{E}}$.
  \item \textbf{Full Factor Sampling and Fresh Contiguous Block Re-Masking:}
  We sample a complete clean discrete factor hypothesis across all $2k$ factor
  positions simultaneously:
  $\hat{x}_{0, j}^{(t)} \sim \mathrm{Categorical}(\tilde{\boldsymbol{\ell}}_j^{(t)})$
  for $j \in \{2k+1, \dots, 4k\}$ (keeping $\hat{x}_{0, 1:2k}^{(t)} = x_{1:2k}$).
  If $t = T - 1$ ($\eta_T = 0$), the sampler returns $\hat{\mathbf{x}}_0^{(T-1)}$
  without re-masking. Otherwise ($t < T - 1$), rather than keeping previously
  unmasked bits frozen, the sampler applies the forward contiguous block-masking
  kernel at target noise level $\eta_{t+1}$ to $\hat{\mathbf{x}}_0^{(t)}$.
  Specifically, at inference time we use deterministic floor rounding with a
  single-bit minimum floor:
  \begin{equation}
    M_t = \max\!\big(\lfloor \eta_{t+1} k \rfloor, \, 1\big),
    \label{eq:inference_mask_width}
  \end{equation}
  draw two fresh, independent uniform block offsets
  $s_{t,p}, s_{t,q} \sim \mathcal{U}\{0, \dots, k - M_t\}$, and replace the
  contiguous spans $[2k + 1 + s_{t,p}, \, 2k + s_{t,p} + M_t]$ in $p$ and
  $[3k + 1 + s_{t,q}, \, 3k + s_{t,q} + M_t]$ in $q$ of $\hat{\mathbf{x}}_0^{(t)}$
  with the mask token to form $\tilde{\mathbf{x}}^{(t+1)}$.
\end{enumerate}

\paragraph{Why Scaling $T \gg 2k$ Improves Accuracy Without Intermediate Verification.}
In standard absorptive discrete diffusion where one masked token is permanently
unmasked per step, the process terminates after $2k$ steps unless an external
verifier restarts failed trajectories. By contrast, in our non-monotonic
predict-and-remask sampler, no arithmetic check $\hat{p} \cdot \hat{q} = n$ is
ever performed during the $T$ steps. When $T \gg k$ (e.g., $T = 16{,}384$ for
$k = 15$), the deterministic block width $M_t$ in Eq.~\eqref{eq:inference_mask_width}
remains constant for $\approx T/k$ consecutive steps before decrementing by $1$.
Across those $T/k$ steps, the masked window of width $M_t$ slides to newly
sampled random offsets $s_{t,p}, s_{t,q}$ at every step while the continuous
embedding $\mathbf{S}^{(t+1)}$ carries forward the full probability distribution
over $\{0, 1\}$ at all $2k$ factor bits. Consequently, any factor bit that was
guessed inconsistently at an earlier step is repeatedly re-masked and re-evaluated
in the context of updated beliefs elsewhere on the canvas. In particular, during
the final $T/k$ steps where $0 < \eta_{t+1} < 1/k$, Eq.~\eqref{eq:inference_mask_width}
clamps $M_t = 1$, so the sampler executes $T/k$ steps of single-bit random-scan
blocked-Gibbs refinement over $p$ and $q$ before final readout at $t = T-1$.

\subsubsection{Outer MSB-to-LSB Beam Search with Inner Mini-Denoising}
\label{app:outer_inner_beam}

While non-monotonic ancestral sampling refines a single trajectory, it can still
be trapped in local modes when early high-order bit guesses are mutually
inconsistent with $n$. Guided beam search replaces single-trajectory sampling
with a two-loop algorithm: an \textbf{outer tree-search loop} over $i = 0, 1, \dots, 2k-1$
that permanently commits one factor bit per step from MSB
to LSB, and an \textbf{inner mini-denoising loop} of
$T_{\text{mini}} = 5$ recurrent self-conditioning steps that estimates calibrated
branching probabilities for the next undecided bit.

\paragraph{Alternating MSB-to-LSB Expansion Schedule and Frontier Initialization.}
Let $p = (p_0, p_1, \dots, p_{k-1})$ and $q = (q_0, q_1, \dots, q_{k-1})$ denote
the $k$ bits of each factor ordered from MSB ($r = 0$, weight $2^{k-1}$) to LSB
($r = k-1$, weight $2^0$), occupying canvas indices $2k+1+r$ and $3k+1+r$,
respectively. The outer loop alternates between factors at each depth
$i \in \{0, \dots, 2k-1\}$:
\begin{equation}
  t_i =
  \begin{cases}
    2k + 1 + \lfloor i/2 \rfloor & \text{if } i \text{ is even (deciding MSB } p_{\lfloor i/2 \rfloor}\text{)}, \\
    3k + 1 + \lfloor i/2 \rfloor & \text{if } i \text{ is odd (deciding MSB } q_{\lfloor i/2 \rfloor}\text{)}.
  \end{cases}
  \label{eq:msb_schedule}
\end{equation}
After $i$ outer steps, every node in the beam shares the same set of decided
canvas positions $\mathcal{D}_i = \{t_0, t_1, \dots, t_{i-1}\}$ ($|\mathcal{D}_i| = i$)
and undecided masked positions
$\mathcal{U}_i = \{2k+1, \dots, 4k\} \setminus \mathcal{D}_i$ ($|\mathcal{U}_i| = 2k - i$).

To maintain fixed tensor shapes of batch size $B$ on accelerators without
evaluating duplicate root nodes at early depths, the frontier at $i = 0$ is
initialized with $B$ slots, $(\tilde{\mathbf{x}}^{[b]}, \text{score}^{[b]}, \text{valid}^{[b]})$ for $b \in \{0, \dots, B-1\}$,
where every slot contains $n$ followed by $2k$ mask tokens, but \emph{only slot $0$
is marked active}:
\begin{equation}
  \text{score}^{[0]} = 0.0, \;\; \text{valid}^{[0]} = \text{True},
  \qquad
  \text{score}^{[b]} = -\infty, \;\; \text{valid}^{[b]} = \text{False} \quad \forall\, b \in \{1, \dots, B-1\}.
\end{equation}
Because invalid slots carry score $-\infty$, their children are automatically
discarded during top-$B$ selection. Thus, at $i=0$ only slot $0$ expands into $2$
valid children ($p_0 \in \{0, 1\}$); at $i=1$ those $\le 2$ valid states expand
into $\le 4$ valid children, and the active frontier grows as
$1 \to 2 \to 4 \to \dots \to B$.

\paragraph{Inner Mini-Denoising Loop ($T_{\text{mini}} = 5$): Frozen Canvas, Recurrent Self-Conditioning.}
At outer step $i$, before branching on position $t_i$, we run an inner loop of
$T_{\text{mini}} = 5$ steps ($m = 0, 1, \dots, T_{\text{mini}}-1$) on each beam
slot's partial canvas $\tilde{\mathbf{x}}$. In sharp contrast to ancestral
sampling, \textbf{there is no re-masking or token sampling inside
the inner loop}:
\begin{itemize}
  \item The discrete input canvas $\tilde{\mathbf{x}}$ remains \emph{completely
  frozen} across all $m \in \{0, \dots, T_{\text{mini}}-1\}$: positions
  $j \in \mathcal{D}_i$ hold their committed bits $x_j^* \in \{0, 1\}$, and
  positions $u \in \mathcal{U}_i$ remain set to the mask token. Thus, the
  discrete fraction of masked factor tokens is constant at $\eta_i = \frac{2k - i}{2k}$.
  \item At $m = 0$, the inner rollout initializes $\mathbf{S}^{(0)} = \mathbf{0}$
  and evaluates the denoiser to obtain initial logits $\boldsymbol{\ell}^{(0)}$.
  \item For each subsequent inner step $m \in \{0, \dots, T_{\text{mini}}-2\}$,
  the model constructs the \emph{oracle self-conditioning} tensor
  $\mathbf{S}^{\text{oracle},(m+1)} \in \mathbb{R}^{4k \times d}$ by clamping all
  search-decided positions $j \in \mathcal{D}_i$ to their exact one-hot token
  embeddings while feeding back the temperature-scaled soft binary predictions at
  undecided positions $u \in \mathcal{U}_i$:
  \begin{equation}
    \mathbf{S}_j^{\text{oracle},(m+1)} =
    \begin{cases}
      \mathbf{0} & \text{if } 1 \le j \le 2k \quad \text{(semiprime prefix } n\text{)}, \\[4pt]
      \sqrt{d} \, \mathbf{e}_{x_j^*}^\top \hat{\mathbf{E}}
      & \text{if } j \in \mathcal{D}_i \quad \text{(search-decided bits, one-hot)}, \\[4pt]
      \sqrt{d} \, \mathrm{softmax}(\tilde{\boldsymbol{\ell}}_j^{(m)}) \hat{\mathbf{E}}
      & \text{if } j \in \mathcal{U}_i \quad \text{(undecided bits, soft logits)}.
    \end{cases}
    \label{eq:oracle_sc}
  \end{equation}
  \item After $T_{\text{mini}}$ forward passes, we read out the binary
  log-probabilities $\log P(x_{t_i} = v \mid \mathcal{D}_i) = \big(\log\mathrm{softmax}(\boldsymbol{\ell}_{t_i}^{(T_{\text{mini}}-1)})\big)_v$
  for $v \in \{0, 1\}$ solely at the next target MSB position $t_i$.
\end{itemize}

\paragraph{Why Clamping Decided Positions in Oracle Self-Conditioning Matters.}
A natural question is why Eq.~\eqref{eq:oracle_sc} must explicitly inject the
non-zero one-hot embedding $\sqrt{d}\,\mathbf{e}_{x_j^*}^\top\hat{\mathbf{E}}$ at
already-decided positions $j \in \mathcal{D}_i$, given that (i) the discrete
canvas $\tilde{\mathbf{x}}$ already contains $x_j^*$ at position $j$, and (ii)
the output logits at $j \in \mathcal{D}_i$ are discarded after the inner loop.
The reason is that $\mathbf{S}^{\text{oracle},(m)}$ modifies the \emph{input} to
layer~0 of the bidirectional transformer via
$\mathbf{Z}_j^{(0)} = \mathrm{RMSNorm}\big(\mathbf{C}_j + \mathbf{G}_{\text{SC}}(\mathbf{S}_j^{\text{oracle},(m)})\big)$
(Eq.~\ref{eq:sc_combine}). Because every transformer block applies bidirectional
all-to-all self-attention across all $4k$ positions, every undecided masked
position $u \in \mathcal{U}_i$ attends directly to the keys and values produced
at the decided positions $j \in \mathcal{D}_i$:
\begin{enumerate}
  \item \textbf{Matching the Pass-2 Training Distribution:} During Pass~2 of
  training (Appendix~\ref{app:training_rollout_and_loss}), unmasked factor
  positions $j \in \{2k+1, \dots, 4k\}$ carry both their discrete token embedding
  $\mathbf{C}_j = \sqrt{d}\,\mathbf{e}_{x_j}^\top\hat{\mathbf{E}}$ and a sharp,
  near-one-hot self-conditioning vector $\mathbf{G}_{\text{SC}}(\mathbf{S}_j^{(1)}) \approx \mathbf{G}_{\text{SC}}(\sqrt{d}\,\mathbf{e}_{x_j}^\top\hat{\mathbf{E}})$.
  If we instead passed $\mathbf{S}_j = \mathbf{0}$ at decided positions
  $j \in \mathcal{D}_i$ for $m \ge 1$, the layer-0 representation at $j \in \mathcal{D}_i$
  would remain $\mathrm{RMSNorm}(\mathbf{C}_j + \mathbf{0})$ (a Pass-1
  representation) while undecided positions $u \in \mathcal{U}_i$ would carry
  Pass-2 representations $\mathrm{RMSNorm}(\mathbf{C}_{\text{mask}} + \mathbf{G}_{\text{SC}}(\mathbf{S}_u^{(m)}))$,
  creating a distribution mismatch across the attention context.
  \item \textbf{Enforcing Branch Consistency on Lower-Probability Beams:} When
  beam search explores a branch that committed to a lower-probability bit
  assignment $x_j^*$ (e.g., $\hat{P}(x_j = 0) = 0.2$, so the branch committed to
  $x_j^* = 0$ while the model's unconditioned Pass-1 logit at position $j$ still
  favours $1$ with probability $0.8$), standard soft self-conditioning would feed
  a vector weighted $80\%$ toward ``$1$'' into $\mathbf{G}_{\text{SC}}(\mathbf{S}_j)$,
  directly contradicting the discrete canvas token $\mathbf{C}_j$ (``$0$'').
  Clamping $\mathbf{S}_j^{\text{oracle}}$ to the exact one-hot embedding of
  $x_j^*$ forces both the canvas pathway $\mathbf{C}_j$ and the self-conditioning
  pathway $\mathbf{G}_{\text{SC}}(\mathbf{S}_j^{\text{oracle}})$ to agree
  unanimously on the branch commitment $x_j^*$.
\end{enumerate}

\paragraph{Arithmetic Interval Feasibility Pruning.}
When each active beam hypothesis $b$ expands at position $t_i$ into two children
$c \in \{2b, 2b+1\}$ assigning $x_{t_i}^* = 0$ and $x_{t_i}^* = 1$ with updated
cumulative log-probability $\text{score}_{\text{child}} = \text{score}^{[b]} + \log P(x_{t_i} = x_{t_i}^* \mid \mathcal{D}_i)$,
the committed MSB prefixes define closed integer intervals $[p_{\min}, p_{\max}]$
and $[q_{\min}, q_{\max}]$ for the two factors:
\begin{align}
  p_{\min} &= \sum_{r \,:\, 2k+1+r \in \mathcal{D}_{i+1}} x_{2k+1+r}^* \, 2^{k-1-r},
  &
  p_{\max} &= p_{\min} + \sum_{r \,:\, 2k+1+r \in \mathcal{U}_{i+1}} 2^{k-1-r},
  \label{eq:p_bounds} \\
  q_{\min} &= \sum_{r \,:\, 3k+1+r \in \mathcal{D}_{i+1}} x_{3k+1+r}^* \, 2^{k-1-r},
  &
  q_{\max} &= q_{\min} + \sum_{r \,:\, 3k+1+r \in \mathcal{U}_{i+1}} 2^{k-1-r}.
  \label{eq:q_bounds}
\end{align}
Because $p, q \ge 0$, any completion of the undecided bits in $\mathcal{U}_{i+1}$
must yield a product in $[p_{\min} q_{\min}, \, p_{\max} q_{\max}]$. Thus, a child
hypothesis is arithmetically feasible if and only if
\begin{equation}
  p_{\min} \cdot q_{\min} \;\le\; n \;\le\; p_{\max} \cdot q_{\max}.
  \label{eq:interval_feasibility}
\end{equation}
Any child violating Eq.~\eqref{eq:interval_feasibility} is immediately pruned by
setting $\text{score}_{\text{child}} = -\infty$ and marking $\text{valid}_{\text{child}} = \text{False}$
before selecting the top $B$ children for outer step $i+1$.

\subsubsection{Worked Example Walkthrough of Beam Search and Oracle Self-Conditioning}
\label{app:worked_example}

To make the interplay between the outer search loop, the $T_{\text{mini}}$-step
inner loop, oracle self-conditioning, and arithmetic interval pruning concrete,
consider factoring the $8$-bit semiprime $n = 143 = 13 \times 11$ with $k = 4$
factor bits ($p = 13 = 1101_2$, $q = 11 = 1011_2$, and $n = 10001111_2$), beam width
$B = 2$, and $T_{\text{mini}} = 5$. The canvas has length $L = 4k = 16$:
\begin{equation*}
  \mathbf{x} = \big(\underbrace{1,0,0,0,1,1,1,1}_{n = 143\text{ (indices }1\dots 8\text{)}} \;\|\;
  \underbrace{p_0, p_1, p_2, p_3}_{p\text{ (indices }9\dots 12\text{)}} \;\|\;
  \underbrace{q_0, q_1, q_2, q_3}_{q\text{ (indices }13\dots 16\text{)}}\big),
\end{equation*}
where $p_0, q_0$ have bit value $2^3 = 8$, $p_1, q_1$ have value $2^2 = 4$,
$p_2, q_2$ have value $2^1 = 2$, and $p_3, q_3$ have value $2^0 = 1$.

\paragraph{Initialization ($i = 0$).}
The frontier has $B = 2$ slots with $\mathcal{D}_0 = \emptyset$ and all $8$
factor bits masked (\texttt{M}):
\begin{itemize}
  \item \textbf{Slot 0:} $\tilde{\mathbf{x}}^{[0]} = [10001111 \;\|\; \text{M M M M} \;\|\; \text{M M M M}]$, $\text{score}^{[0]} = 0.0$, $\text{valid}^{[0]} = \text{True}$.
  \item \textbf{Slot 1:} $\tilde{\mathbf{x}}^{[1]} = [10001111 \;\|\; \text{M M M M} \;\|\; \text{M M M M}]$, $\text{score}^{[1]} = -\infty$, $\text{valid}^{[1]} = \text{False}$.
\end{itemize}

\paragraph{Outer Step $i = 0$ (Deciding MSB $p_0$ at Canvas Index $t_0 = 9$).}
\begin{enumerate}
  \item \emph{Inner Mini-Denoising ($m = 0, \dots, 4$):} Throughout all $5$ inner
  passes, the discrete canvas $\tilde{\mathbf{x}}^{[0]}$ stays frozen with all $8$
  factor positions masked. At $m = 0$, $\mathbf{S}^{(0)} = \mathbf{0}$, producing
  initial soft probabilities (say $\hat{P}^{(0)} = [0.75, 0.52, 0.40, 0.90 \;\|\; 0.72, 0.48, 0.55, 0.90]$).
  Since $\mathcal{D}_0 = \emptyset$, for $m = 1, 2, 3, 4$ the oracle
  self-conditioning tensor $\mathbf{S}^{\text{oracle},(m)}$ simply feeds back the
  evolving soft embeddings $\sqrt{d}\,\mathrm{softmax}(\tilde{\boldsymbol{\ell}}_u^{(m-1)})\hat{\mathbf{E}}$
  at all $8$ masked factor positions, sharpening the marginal belief at $p_0$ to,
  say, $P^{(4)}(p_0 = 1) = 0.92$ and $P^{(4)}(p_0 = 0) = 0.08$.
  \item \emph{Branching and Feasibility Pruning:} Expanding Slot~0 on $p_0 \in \{0, 1\}$
  (while Slot~1's children inherit score $-\infty$) yields two active candidate
  children with $\mathcal{D}_1 = \{9\}$:
  \begin{itemize}
    \item \textbf{Child A ($p_0^* = 0$):} Canvas $[10001111 \;\|\; \mathbf{0}\text{ M M M} \;\|\; \text{M M M M}]$, score $\log(0.08) = -2.53$. By Eqs.~\eqref{eq:p_bounds}--\eqref{eq:q_bounds}, $p \in [0000_2, 0111_2] = [0, 7]$ and $q \in [0000_2, 1111_2] = [0, 15]$, so the maximum possible product is $p_{\max} q_{\max} = 7 \times 15 = 105 < 143$. Because $143 \notin [0, 105]$, Child~A is \textbf{pruned immediately} ($\text{score} \leftarrow -\infty, \text{valid} \leftarrow \text{False}$).
    \item \textbf{Child B ($p_0^* = 1$):} Canvas $[10001111 \;\|\; \mathbf{1}\text{ M M M} \;\|\; \text{M M M M}]$, score $\log(0.92) = -0.083$. Here $p \in [1000_2, 1111_2] = [8, 15]$ and $q \in [0, 15]$, giving feasible product interval $[0, 225] \ni 143$ (\textbf{valid}).
  \end{itemize}
  \item \emph{Top-$2$ Selection:} Slot~0 becomes $[10001111 \;\|\; \mathbf{1}\text{ M M M} \;\|\; \text{M M M M}]$ ($\text{score} = -0.083, \text{valid} = \text{True}$), and Slot~1 remains invalid ($\text{score} = -\infty$).
\end{enumerate}

\paragraph{Outer Step $i = 1$ (Deciding MSB $q_0$ at Canvas Index $t_1 = 13$).}
\begin{enumerate}
  \item \emph{Inner Mini-Denoising with Oracle Self-Conditioning ($m = 0, \dots, 4$):}
  Now $\mathcal{D}_1 = \{9\}$ ($p_0^* = 1$) and $\mathcal{U}_1 = \{10, 11, 12, 13, 14, 15, 16\}$.
  On every inner step $m \ge 1$, Eq.~\eqref{eq:oracle_sc} clamps index~$9$ of
  $\mathbf{S}^{\text{oracle},(m)}$ to the exact one-hot embedding
  $\sqrt{d}\,\mathbf{e}_1^\top\hat{\mathbf{E}}$, while indices $10\dots 16$
  receive soft probability-weighted embeddings. After $5$ inner passes, suppose
  $P^{(4)}(q_0 = 1 \mid p_0^* = 1) = 0.94$ and $P^{(4)}(q_0 = 0 \mid p_0^* = 1) = 0.06$.
  \item \emph{Branching and Feasibility Pruning:} Expanding Slot~0 on $q_0 \in \{0, 1\}$:
  \begin{itemize}
    \item \textbf{Child A ($p_0^* = 1, q_0^* = 0$):} $p \in [8, 15]$ and $q \in [0000_2, 0111_2] = [0, 7]$, so $p_{\max} q_{\max} = 15 \times 7 = 105 < 143$ (\textbf{pruned}).
    \item \textbf{Child B ($p_0^* = 1, q_0^* = 1$):} $p \in [8, 15]$ and $q \in [8, 15]$, so $[p_{\min} q_{\min}, p_{\max} q_{\max}] = [64, 225] \ni 143$ (\textbf{valid}, cumulative score $-0.083 + \log(0.94) = -0.145$).
  \end{itemize}
  \item \emph{Top-$2$ Selection:} Slot~0 holds $[10001111 \;\|\; \mathbf{1}\text{ M M M} \;\|\; \mathbf{1}\text{ M M M}]$ ($\text{score} = -0.145, \text{valid} = \text{True}$).
\end{enumerate}

\paragraph{Outer Step $i = 2$ (Deciding Second MSB $p_1$ at Canvas Index $t_2 = 10$).}
\begin{enumerate}
  \item \emph{Inner Mini-Denoising ($m = 0, \dots, 4$):} Now $\mathcal{D}_2 = \{9, 13\}$
  ($p_0^* = 1, q_0^* = 1$) are both clamped to one-hot embeddings in
  $\mathbf{S}^{\text{oracle},(m)}$, while the $6$ undecided bits refine their
  soft embeddings. Suppose after $m = 4$ the model predicts
  $P^{(4)}(p_1 = 1 \mid p_0^* = 1, q_0^* = 1) = 0.65$ and
  $P^{(4)}(p_1 = 0 \mid p_0^* = 1, q_0^* = 1) = 0.35$.
  \item \emph{Branching and Feasibility Pruning:} Expanding Slot~0 on $p_1 \in \{0, 1\}$:
  \begin{itemize}
    \item \textbf{Child A ($p_0^* = 1, p_1^* = 1, q_0^* = 1$):} $p \in [1100_2, 1111_2] = [12, 15]$ and $q \in [1000_2, 1111_2] = [8, 15]$, giving product interval $[96, 225] \ni 143$ (\textbf{valid}, score $-0.145 + \log(0.65) = -0.576$).
    \item \textbf{Child B ($p_0^* = 1, p_1^* = 0, q_0^* = 1$):} $p \in [1000_2, 1011_2] = [8, 11]$ and $q \in [8, 15]$, giving product interval $[64, 165] \ni 143$ (\textbf{also valid}, score $-0.145 + \log(0.35) = -1.195$).
  \end{itemize}
  \item \emph{Top-$2$ Selection:} Both children satisfy Eq.~\eqref{eq:interval_feasibility},
  so for the first time both slots of the $B = 2$ beam become active: Slot~0
  tracks the prefix $(p_{0:1}^* = 11_2, q_0^* = 1_2)$ and Slot~1 tracks the
  alternative prefix $(p_{0:1}^* = 10_2, q_0^* = 1_2)$. Notice how at $i = 3$,
  when evaluating Slot~1 (where $p_1^* = 0$ was chosen against the model's $0.65$
  preference for $p_1 = 1$), clamping index~$10$ of $\mathbf{S}^{\text{oracle},(m)}$
  to the one-hot embedding of ``$0$'' forces the inner mini-denoising loop to
  compute the distribution of $q_1$ conditioned on $p \in [8, 11]$ (which requires
  $q_1 = 1$ since $11 \times 11 = 121 < 143$), rather than leaking the model's
  unconditioned preference for $p_1 = 1$.
\end{enumerate}
Continuing through outer step $i = 7$ ($q_3$), the interval bounds collapse to a
single point $p_{\min} = p_{\max} = \hat{p}$ and $q_{\min} = q_{\max} = \hat{q}$,
where Eq.~\eqref{eq:interval_feasibility} verifies $\hat{p} \cdot \hat{q} = 13 \times 11 = 143$.

\section{Supplementary Power-Law Results}
\label{sec:appendix_scaling}

\subsection{Optimal Compute Allocation}

\begin{figure}[htbp]
    \centering
    \includegraphics[width=\linewidth]{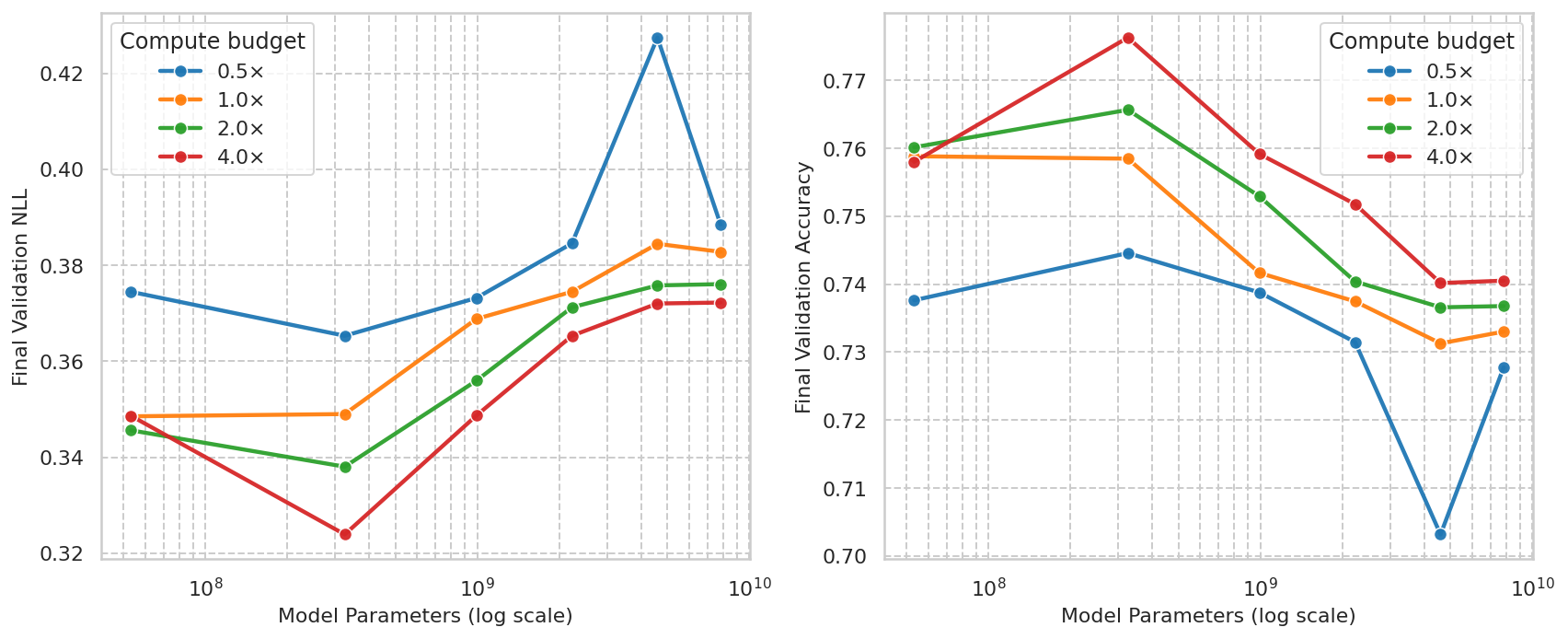}
    \caption{IsoFLOP smile curves for $k=24$, demonstrating the trade-off between final validation NLL (left) and final validation accuracy (right) against model parameters for fixed compute budgets.}
    \label{fig:isoflop}
\end{figure}
To establish the optimal compute allocation in terms of model size and training tokens, we conducted an IsoFLOP analysis for the $k=24$ bit-length factorization task. Figure~\ref{fig:isoflop} shows the characteristic \emph{smile} curves. The minima of these curves correspond to the compute-optimal model size for a given FLOP budget. At these compute-optimal minima ($\approx 3 \times 10^8$ parameters), scaling the FLOP budget from $0.5\times$ to $4.0\times$ pushes validation NLL down from $\approx 0.365$ to $\approx 0.324$ (and validation accuracy up from $\approx 0.745$ to $\approx 0.776$), penetrating well below the unweighted non-factoring local-arithmetic baseline ($\mathrm{NLL} \approx 0.375$, $\mathrm{Accuracy} \approx 0.732$; see Appendix~\ref{app:non_factoring_baseline}). Conversely, on the right-hand branch of the smile curves ($\sim 10^{10}$ parameters), over-allocating FLOPs to model parameters at the expense of training steps leaves the largest models token-starved near the $\mathrm{NLL} \approx 0.375$ / $\mathrm{Accuracy} \approx 0.732$ non-factoring baseline.

\subsection{Impact of Factor Bit-Length}
\label{app:factor_bits_impact}

\begin{figure}[htbp]
    \centering
    \includegraphics[width=\linewidth]{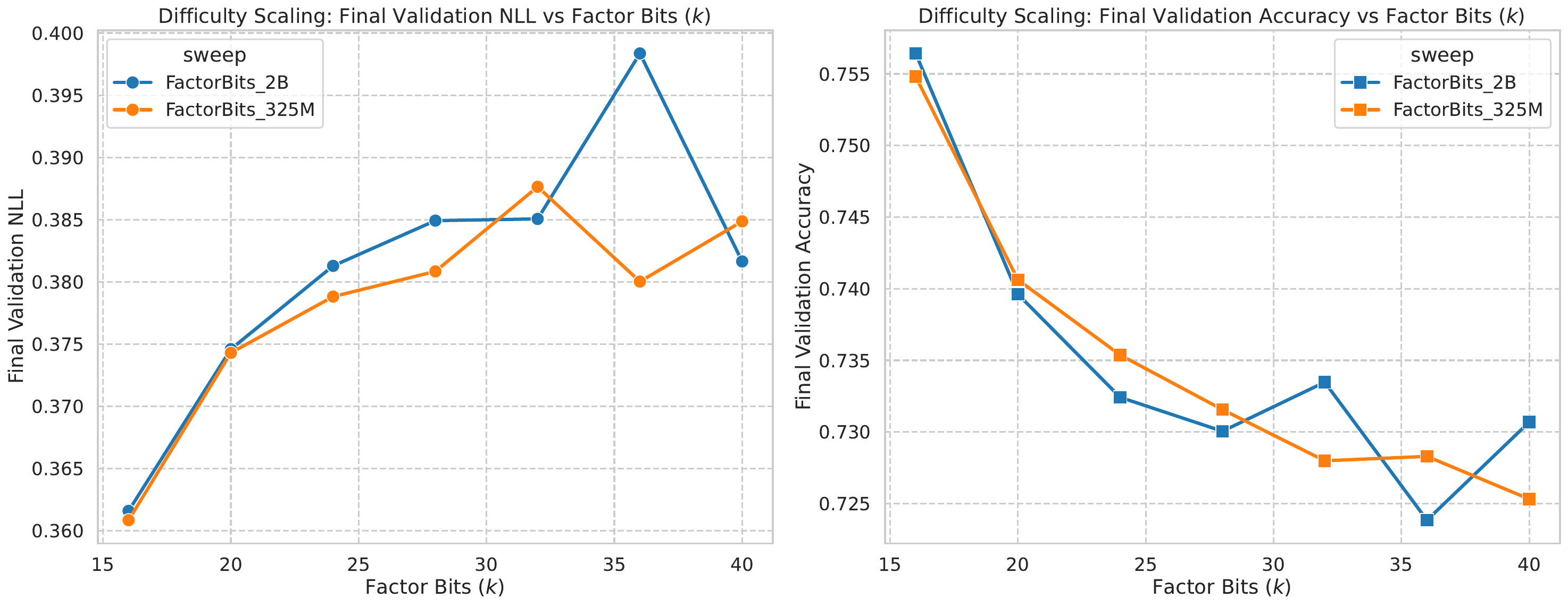}
    \caption{Difficulty scaling illustrating the degradation of final validation NLL (left) and final validation accuracy (right) as the number of factor bits increases, compared between the 325M and 2B models.}
    \label{fig:difficulty_scaling}
\end{figure}

Note on the differences loss scale and step budget vs Figure \ref{fig:scaling_law_model_size}: The main-text trajectories were trained for a fixed 100,000 steps with inverse-PDF noise-level importance weighting, whereas the Appendix sweeps use uniform example weighting across sampled noise levels and scale the training step budget as $2500 \cdot 2^{(k-16)/4}$.

Figure~\ref{fig:difficulty_scaling} illustrates the increasing difficulty of the integer factorization problem as the number of factor bits increases. The final validation NLL increases and the final validation accuracy drops when factoring larger numbers. Because validation NLL and accuracy in Figures~\ref{fig:isoflop}--\ref{fig:training_dynamics_325m_linear} average over all $2k$ factor bits under $\eta \sim 2\eta$ with unit weights $w_b = 1$ (Appendix~\ref{app:non_factoring_baseline}), the plateau at $\mathrm{NLL} \approx 0.38\text{--}0.39$ and $\mathrm{Accuracy} \approx 0.728\text{--}0.734$ for $k \ge 24$ in Figure~\ref{fig:difficulty_scaling} corresponds directly to the non-factoring local-arithmetic baseline of Eq.~\eqref{eq:nll_baseline_unweighted} ($\mathrm{NLL} \approx 0.375$, $\mathrm{Accuracy} \approx 0.732$). At these larger bit-lengths and short step budgets ($2500 \cdot 2^{(k-16)/4}$ steps), the models learn token copying, odd-parity LSBs, and local multiplicative/divisive consistency across partially unmasked factor spans, but make negligible progress on global $\eta \to 1$ factorization; genuine factoring capability emerges only once validation NLL penetrates substantially below this $0.375$ floor (as seen at $k=16$, where NLL reaches $0.360$ and beam search solves $\approx 15.5\%$ of test instances in Figure~\ref{fig:bfs_frontier_by_stage}).

Here we present extended empirical results illustrating the model training dynamics and the power law fits, complementing the results in the main text.

\subsection{Fitting Individual Trajectories}

\begin{figure}[htbp]
    \centering
    \includegraphics[width=\linewidth]{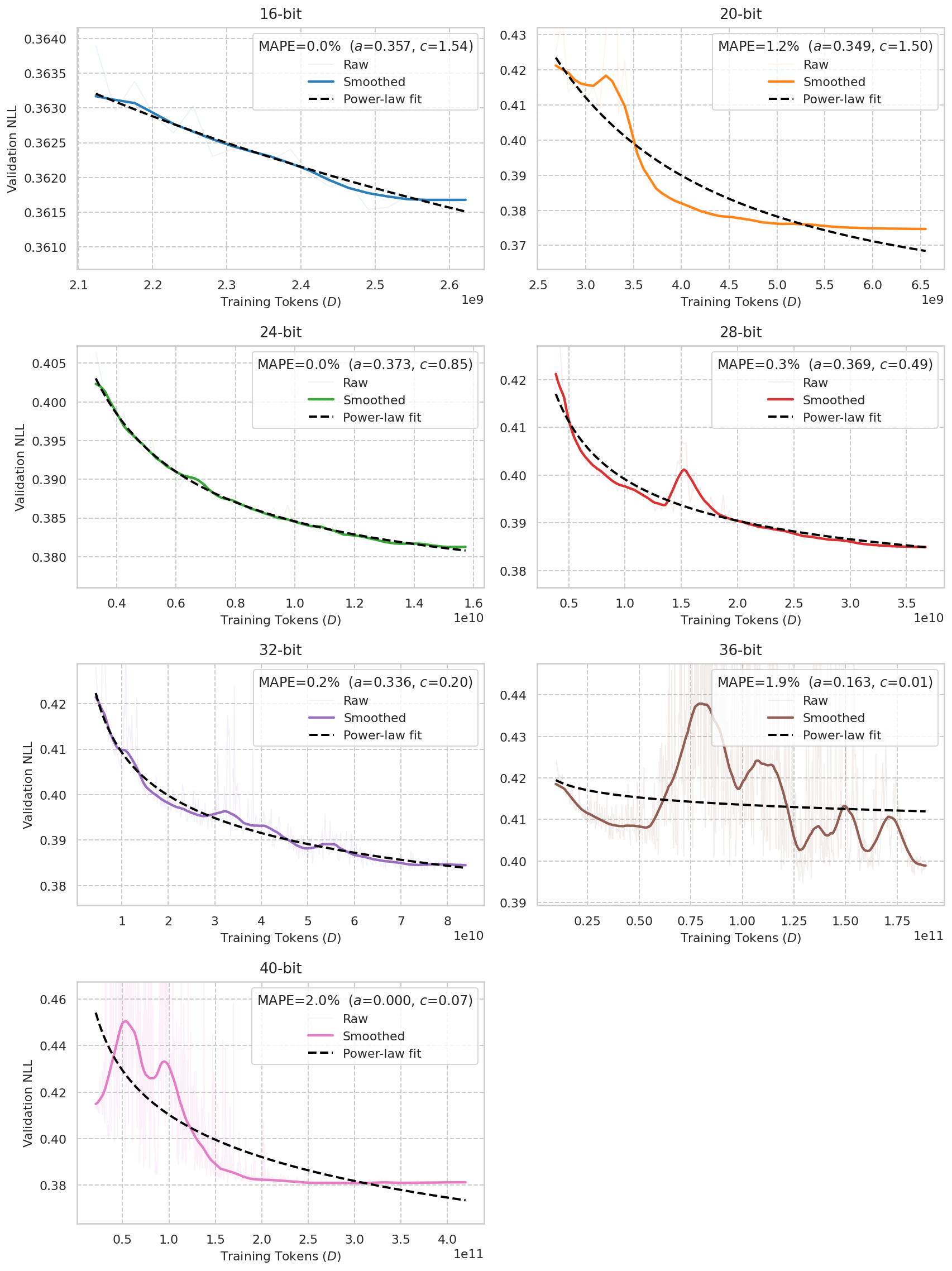}
    \caption{Faceted per-trajectory power-law fits for the 2B parameter model across factor bit-lengths $k=16$ to $k=40$. The fits closely track the empirical validation NLL with low mean absolute percentage error (MAPE).}
    \label{fig:faceted_2b}
\end{figure}

\begin{figure}[htbp]
    \centering
    \includegraphics[width=\linewidth]{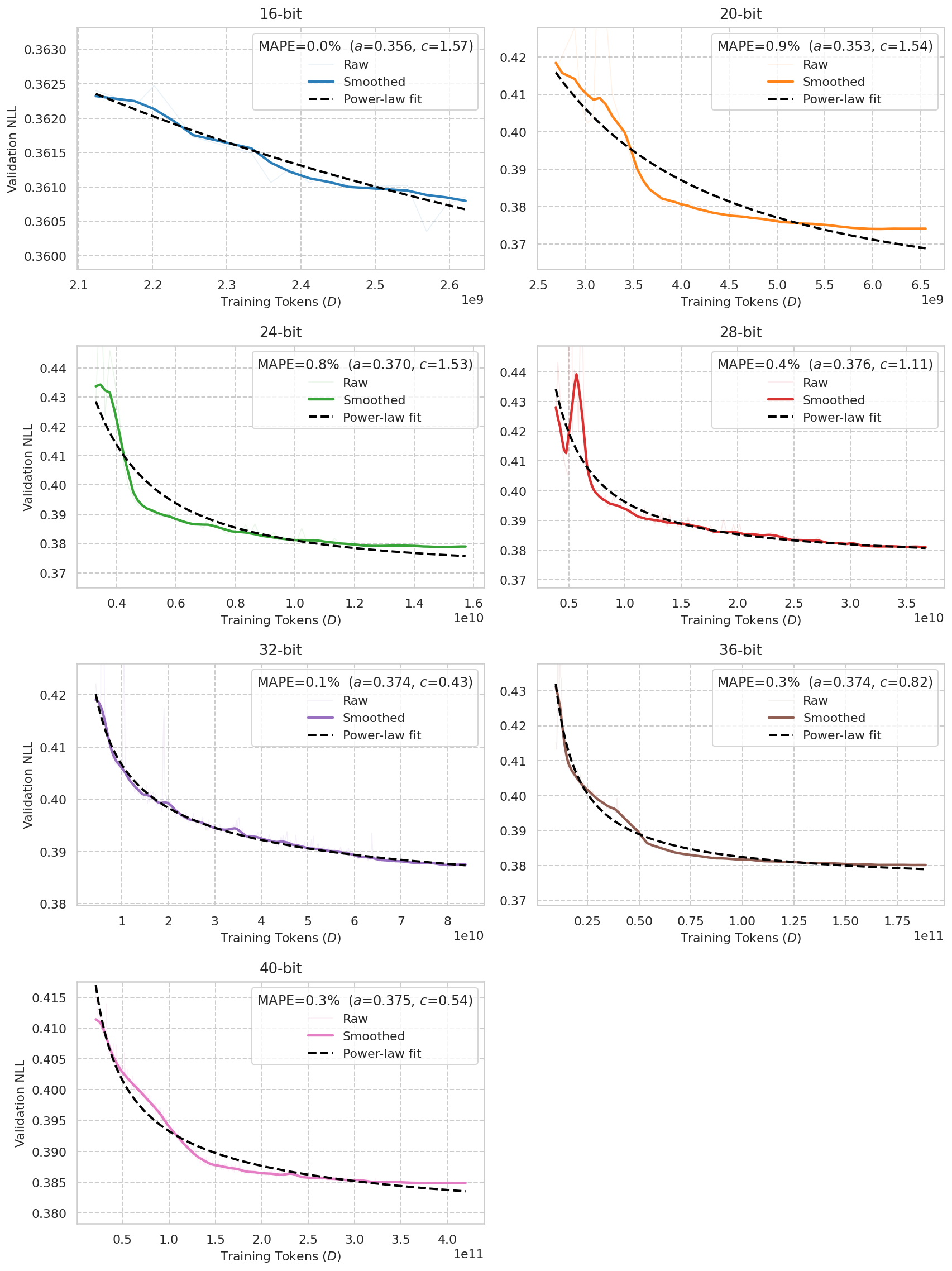}
    \caption{Faceted per-trajectory power-law fits for the 325M parameter model across factor bit-lengths $k=16$ to $k=40$.}
    \label{fig:faceted_325m}
\end{figure}

The faceted per-trajectory power-law fits for the 2B model are shown in Figure~\ref{fig:faceted_2b} and for the 325M model in Figure~\ref{fig:faceted_325m}. The empirical validation NLL is fitted to the functional form $L(D) = a + b \cdot D^{-c}$, where $D$ represents the cumulative training tokens.

The theoretical fit closely matches the empirical data across all evaluated bit-lengths. This indicates that the power-law decay consistently holds at the individual trajectory level. Note that for $k \ge 24$, the fitted asymptotes $a \in [0.370, 0.376]$ (Figure~\ref{fig:faceted_325m}) reflect rapid power-law convergence toward the $\approx 0.375$ non-factoring local-arithmetic floor (Appendix~\ref{app:non_factoring_baseline}) over the short sweep horizon, rather than convergence toward zero-loss factorization.

\subsection{Training Dynamics}

\begin{figure}[htbp!]
    \centering
    \includegraphics[width=\linewidth]{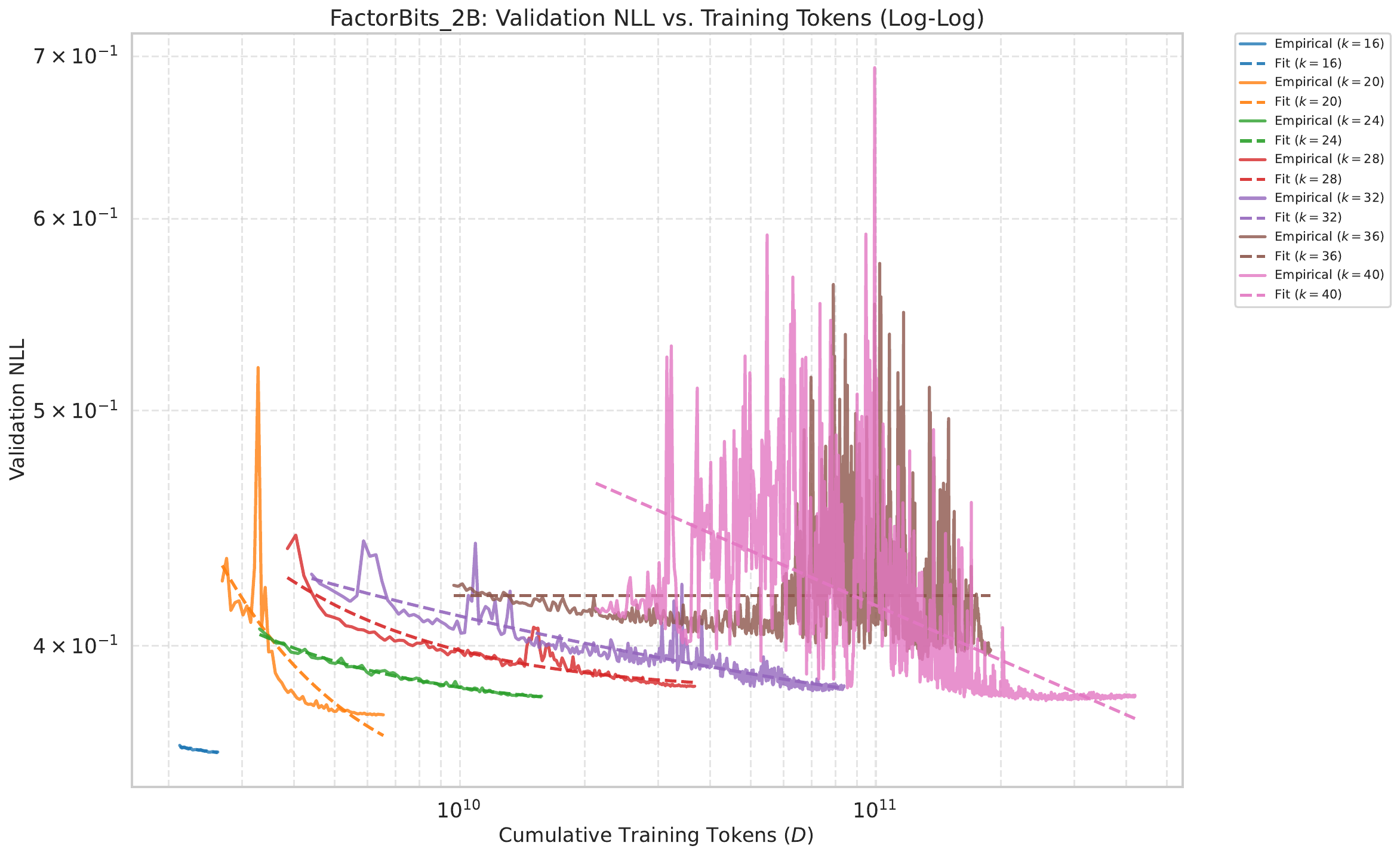}
    \caption{Validation NLL versus cumulative training tokens on a log-log scale for the 2B parameter model across factor bit-lengths $k=16$ to $k=40$.}
    \label{fig:training_dynamics_2B}
\end{figure}

\begin{figure}[htbp!]
    \centering
    \includegraphics[width=\linewidth]{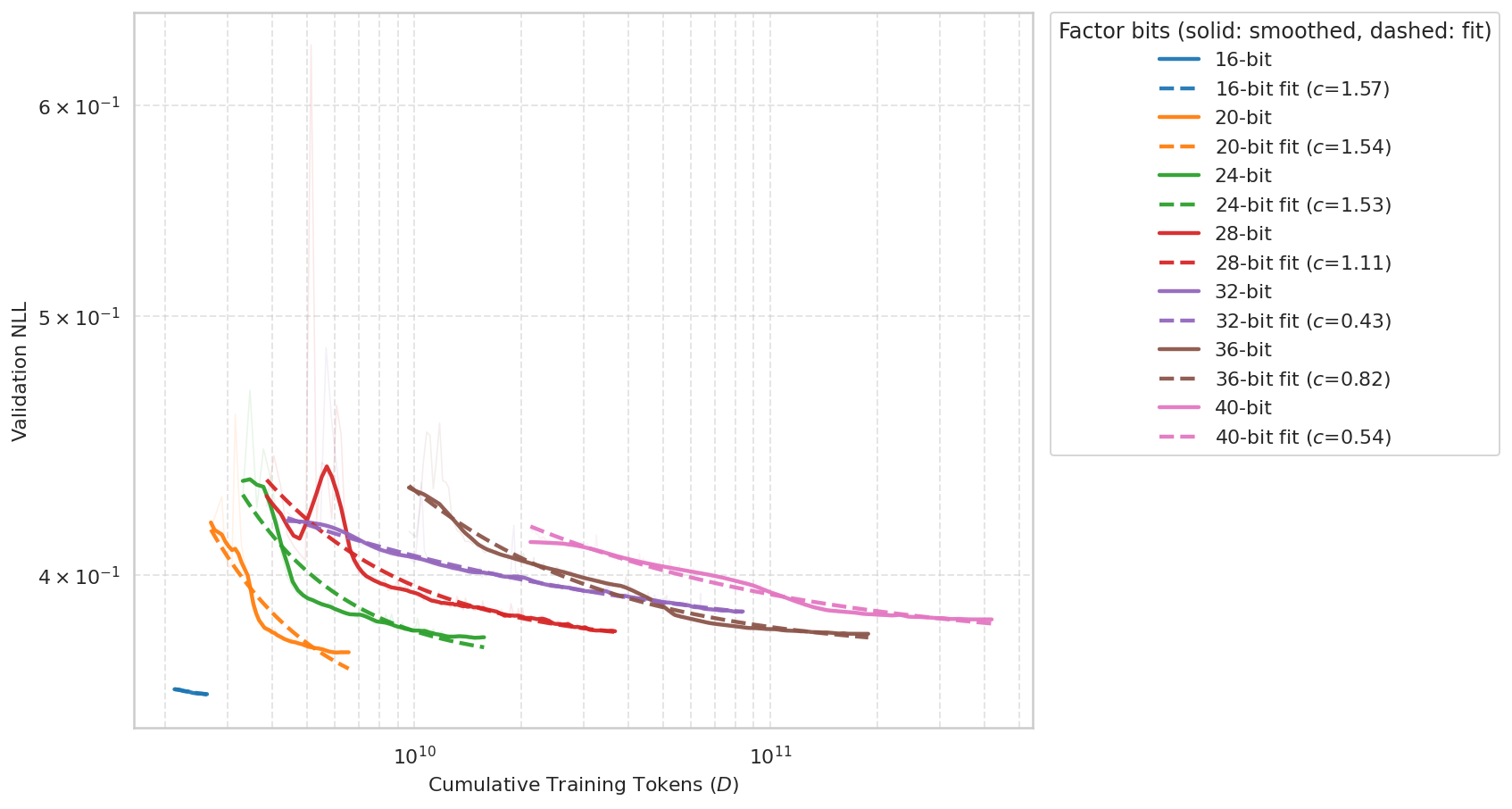}
    \caption{Validation NLL versus training tokens on a log-log scale for the 325M parameter model across various factor bit-lengths.}
    \label{fig:training_dynamics_325m_loglog}
\end{figure}

Figures~\ref{fig:training_dynamics_2B} and~\ref{fig:training_dynamics_325m_loglog} show the log-log scaling trajectories of validation NLL versus cumulative training tokens for the 2B and 325M parameter models across factor bit-lengths $k \in \{16, \dots, 40\}$. The validation NLL decay holds across different model scales, maintaining a linear trend in log-log space.

\begin{figure}[htbp!]
    \centering
    \includegraphics[width=\linewidth]{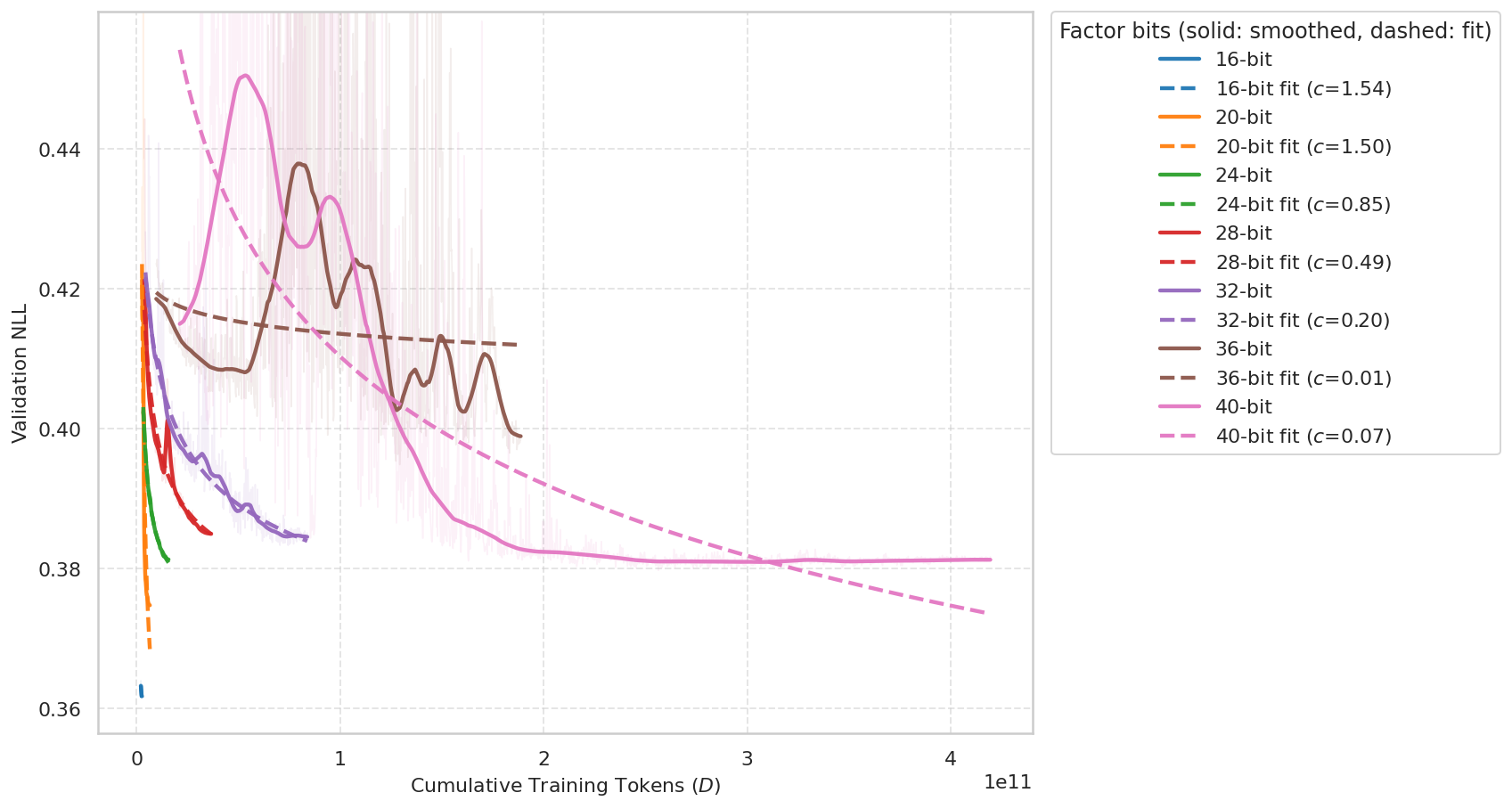}
    \caption{Validation NLL versus cumulative training tokens (linear scale) for the 2B parameter model.}
    \label{fig:training_dynamics_2b_linear}
\end{figure}

\begin{figure}[htbp!]
    \centering
    \includegraphics[width=\linewidth]{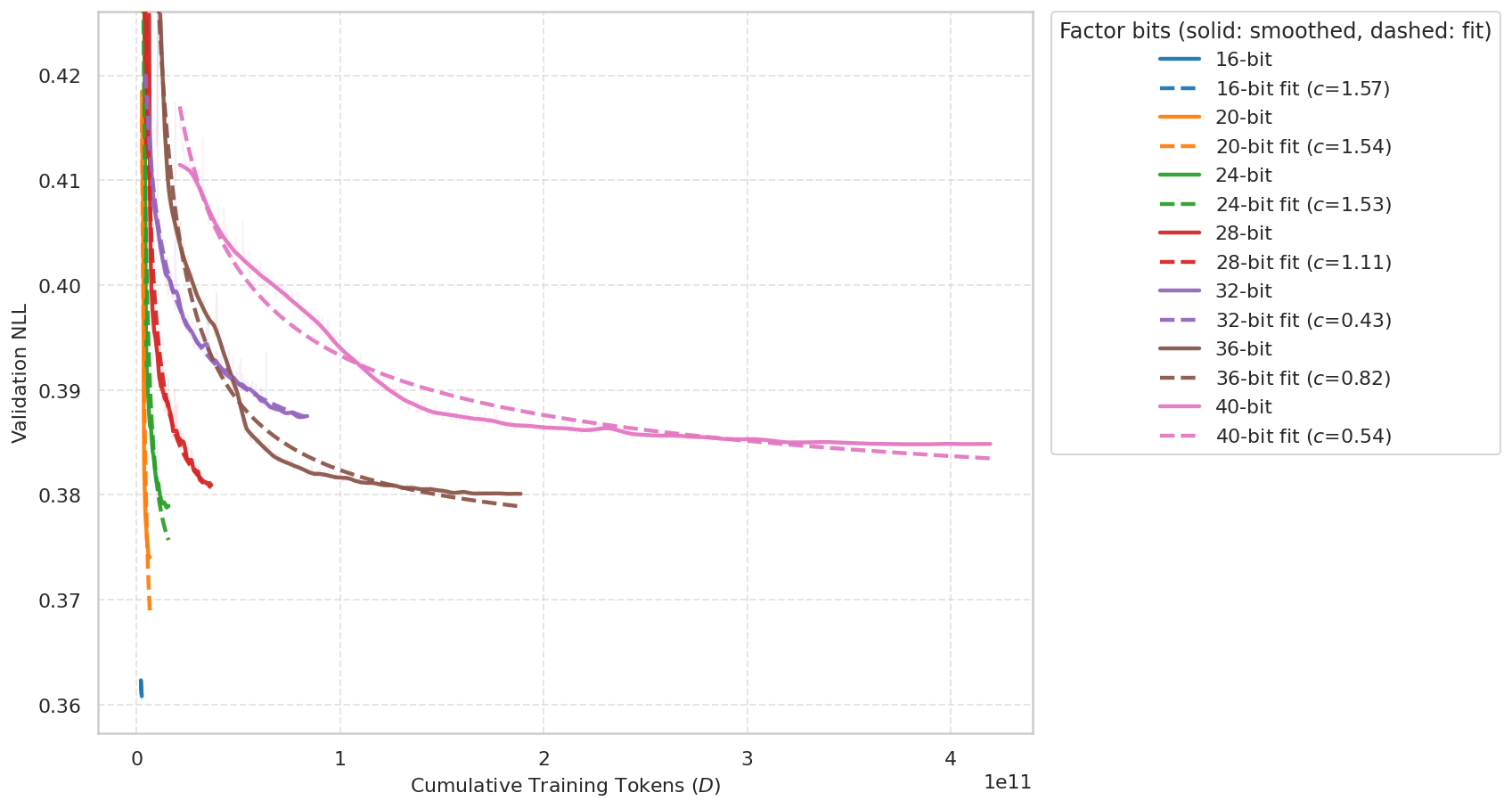}
    \caption{Validation NLL versus cumulative training tokens (linear scale) for the 325M parameter model.}
    \label{fig:training_dynamics_325m_linear}
\end{figure}

Figures~\ref{fig:training_dynamics_2b_linear} and \ref{fig:training_dynamics_325m_linear} present the validation NLL against cumulative training tokens on a standard linear scale for the 2B and 325M models respectively. While these plots clearly show the rapid initial improvement and the long-tail convergence as training progresses, the underlying power-law relationships are heavily compressed. This visualizes why log-log representations are necessary to properly analyze the scaling behavior and extrapolate future performance.

\clearpage
\subsection{Beam Search Frontier Scaling Across Pre-Training Stages}
\label{app:beam_search_scaling}

\begin{figure}[htbp!]
    \centering
    \includegraphics[width=\linewidth]{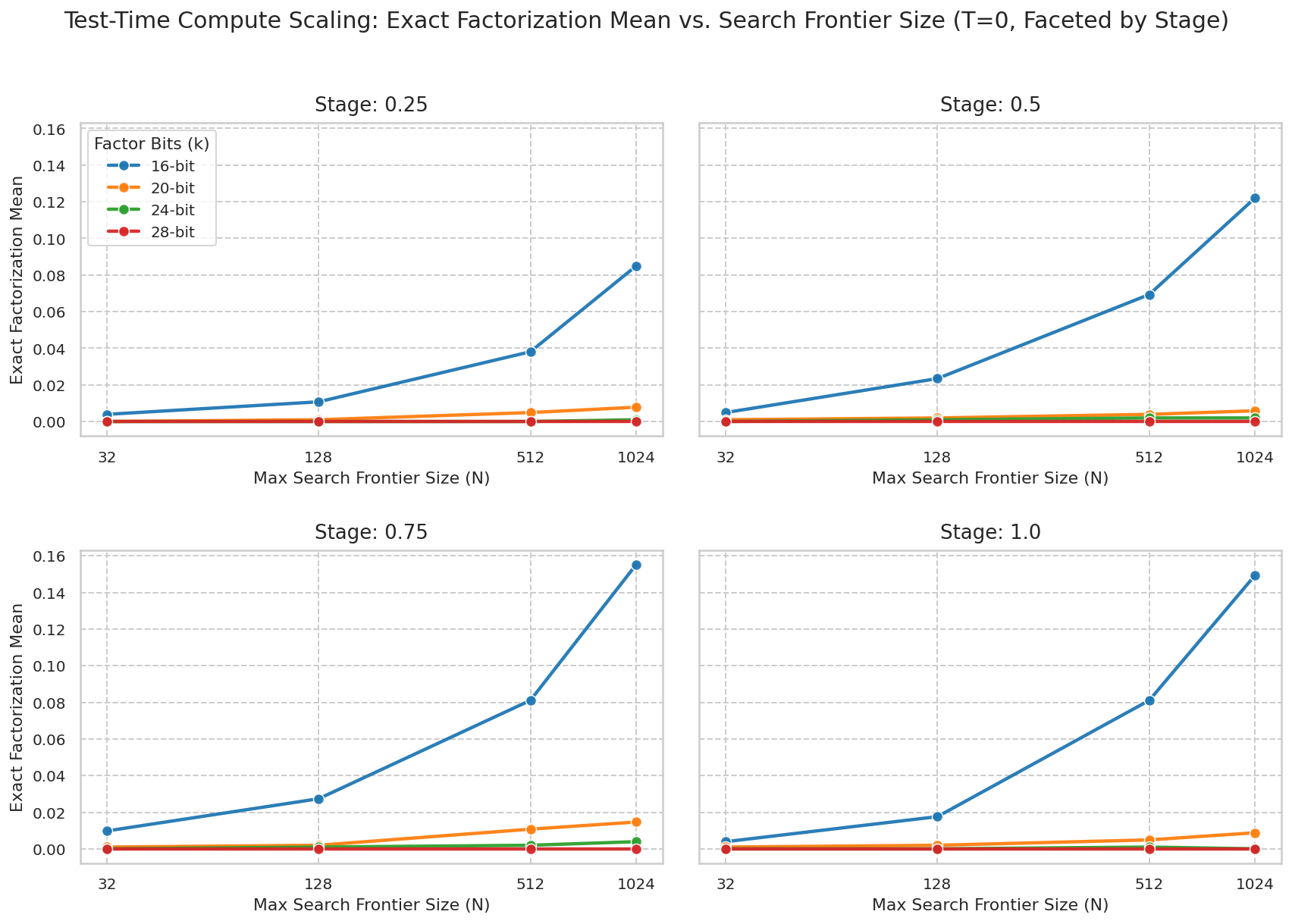}
    \caption{\textbf{Test-Time Compute Scaling across Pre-Training Checkpoints.} Exact factorization mean as a function of max search frontier size ($B \in \{32, 128, 512, 1024\}$, labeled $N$ on the horizontal axis; $\log_2$ scale) across factor bit-lengths ($k \in \{16, 20, 24, 28\}$, greedy decoding temperature $\tau=0$, labeled $T=0$). Subplots denote pre-training progression ($\text{fraction complete} \in \{0.25, 0.50, 0.75, 1.00\}$). Test-time search scaling yields monotonic performance gains, with larger frontier budgets partially compensating for earlier checkpoint maturity.}
    \label{fig:bfs_frontier_by_stage}
\end{figure}

\begin{figure}[htbp!]
    \centering
    \includegraphics[width=\linewidth]{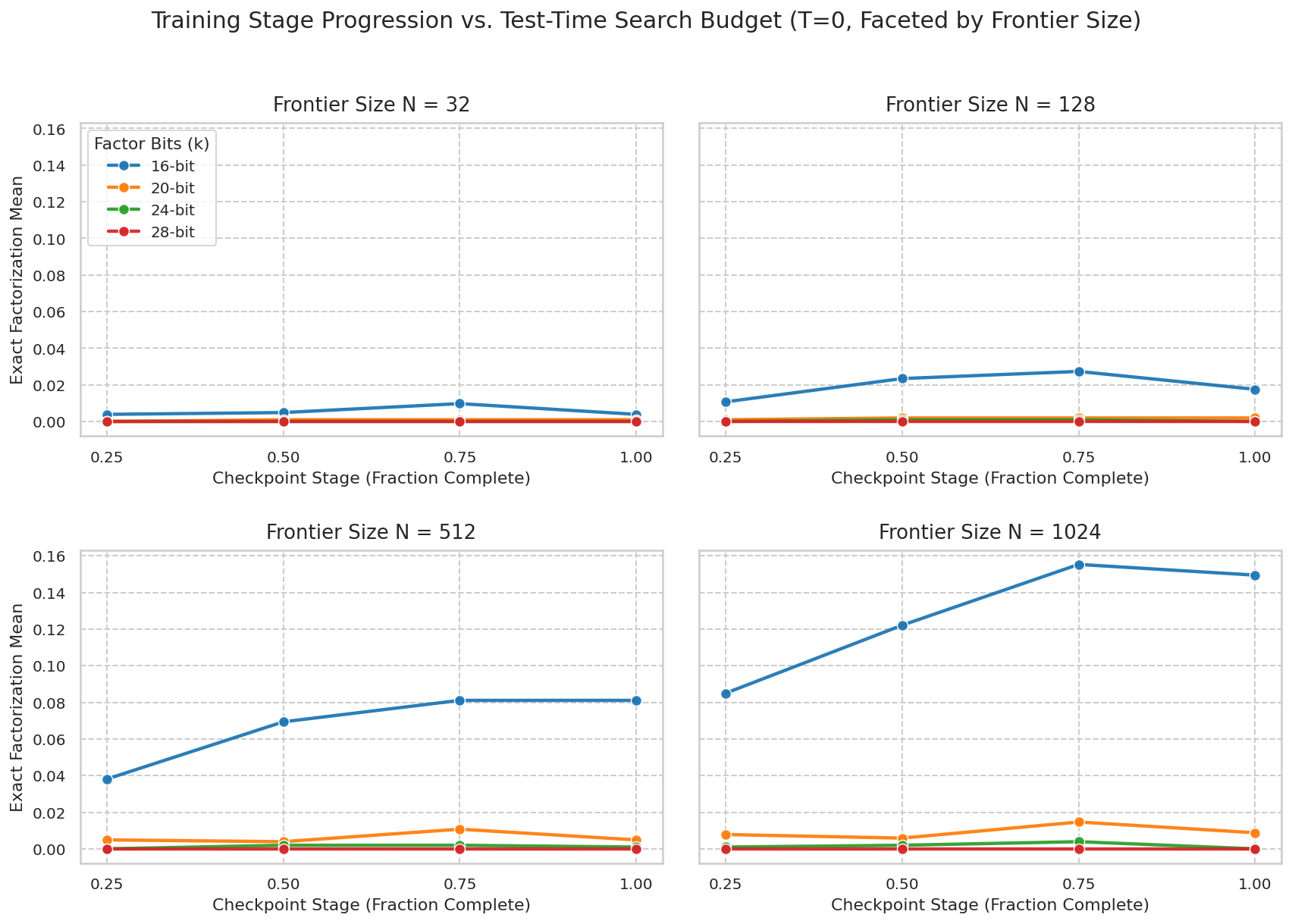}
    \caption{\textbf{Pre-Training Stage Progression vs. Test-Time Search Budget.} Exact factorization mean vs.\ pre-training completion fraction ($0.25$ to $1.00$), faceted by search frontier size ($B \in \{32, 128, 512, 1024\}$, labeled $N$; greedy decoding $\tau=0$). Colored lines denote factor bit-length ($k \in \{16, 20, 24, 28\}$). As pre-training matures, higher search budgets ($B=1024$) unlock substantially steeper performance trajectories compared to constrained search ($B=32$).}
    \label{fig:bfs_stage_progression}
\end{figure}

\begin{figure}[htbp!]
    \centering
    \includegraphics[width=\linewidth]{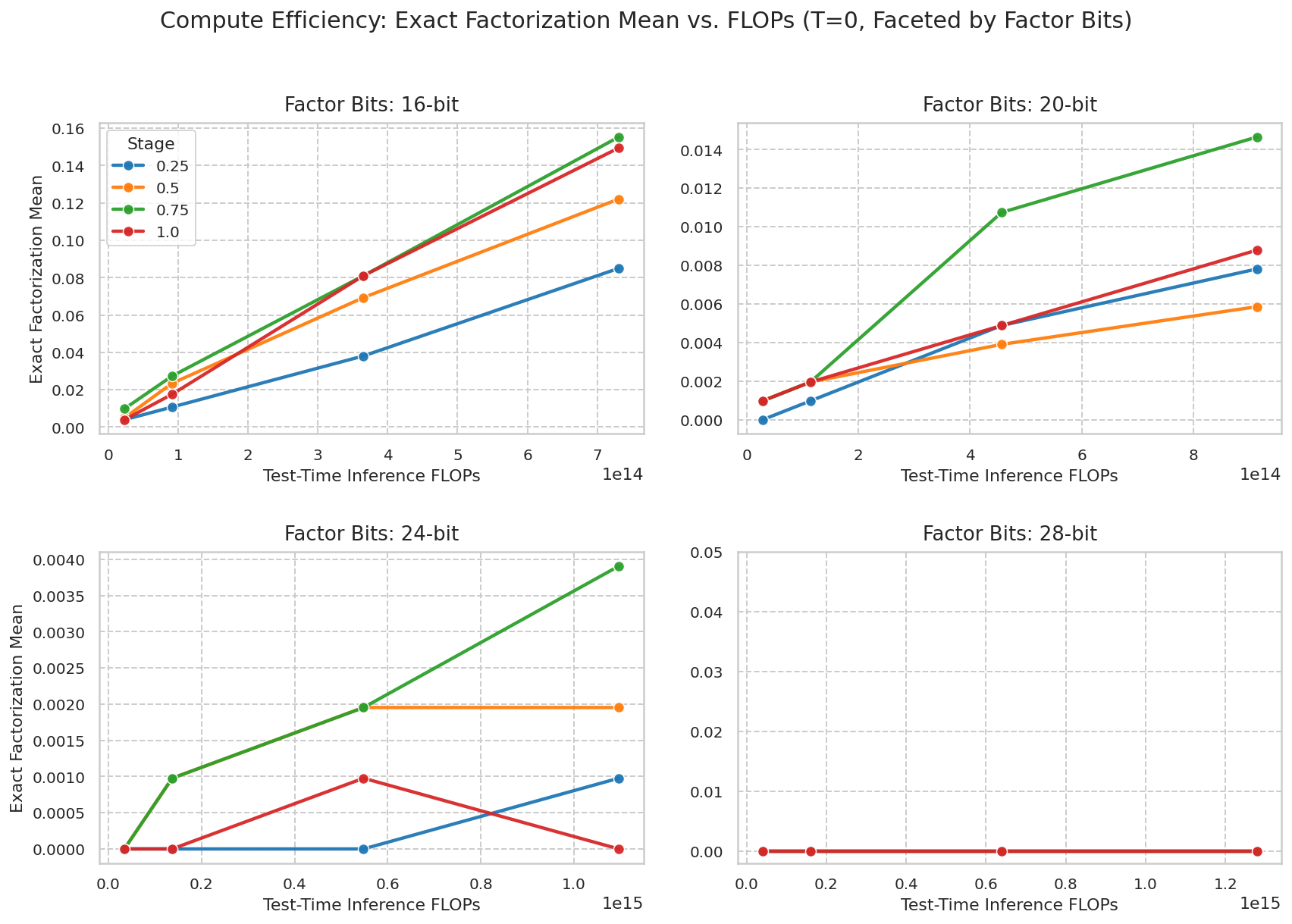}
    \caption{\textbf{Compute Efficiency (Test-Time FLOPs vs. Performance).} Exact factorization mean as a function of total test-time inference FLOPs, faceted by factor bit-length ($k \in \{16, 20, 24, 28\}$ bits, greedy decoding $\tau=0$). Colored curves represent pre-training checkpoint stages ($0.25$ to $1.00$). Note independent $x$- and $y$-axes per subplot. Within each fixed problem hardness, expanding test-time search compute demonstrates predictable efficiency scaling.}
    \label{fig:bfs_flops}
\end{figure}

Figures~\ref{fig:bfs_frontier_by_stage}--\ref{fig:bfs_flops} evaluate beam search with oracle self-conditioning and MSB-to-LSB interval feasibility pruning (Section~\ref{sec:test_time_search}; $T_{\mathrm{mini}} = 5$ inner mini-denoising steps, greedy expansion, and frontier widths $B \in \{32, 128, 512, 1024\}$) across $k \in \{16, 20, 24, 28\}$ on $1{,}024$ held-out test semiprimes per configuration. Unlike Table~\ref{tab:factorization_accuracy}—which evaluates long-horizon checkpoints ($100\text{k}$--$348\text{k}$ training steps) using single-trajectory ancestral sampling—Figures~\ref{fig:bfs_frontier_by_stage}--\ref{fig:bfs_flops} evaluate intermediate and final checkpoints (completion fractions $0.25, 0.50, 0.75, 1.00$) from the short-horizon 2B difficulty-scaling sweep of Appendix~\ref{app:factor_bits_impact}, where total training steps scale as $2500 \cdot 2^{(k-16)/4}$ (i.e., only $2{,}500$ steps at $k=16$ and $5{,}000$ steps at $k=20$). Even on these lightly trained checkpoints (validation NLL $\approx 0.362$ at $k=16$ and $\approx 0.375$ at $k=20$), expanding the search frontier from $B = 32$ to $B = 1024$ increases exact factorization accuracy from $<1\%$ to $\approx 15.5\%$ at $k=16$ and $\approx 1.5\%$ at $k=20$, whereas checkpoints stuck on the non-factoring validation plateau ($k = 28$) remain at $0\%$ across all tested beam widths.

\end{document}